\documentclass[11pt]{article}

\usepackage[final]{acl}

\usepackage{times}
\usepackage{latexsym}

\usepackage[T1]{fontenc}

\usepackage[utf8]{inputenc}

\usepackage{microtype}

\usepackage{inconsolata}

\usepackage{amsmath,amssymb}
\usepackage{booktabs}
\usepackage{url}
\usepackage{enumitem}
\usepackage{caption}
\usepackage{subcaption}
\usepackage{multirow}
\usepackage{float}
\usepackage{placeins} 
\usepackage{graphicx}
\usepackage{hyperref} 
\usepackage{placeins} 
\usepackage{comment}
\usepackage{todonotes}
\usepackage{caption}
\usepackage{float}       
\usepackage{subcaption}  

\title{Beyond Divergent Creativity: A Human-Based Evaluation of Creativity in Large Language Models}

\author{
  Kumiko Nakajima$^{\dagger}$  \quad Jan Zuiderveld$^{\ast}$ \quad Sandro Pezzelle$^{\dagger}$ \\
  $^{\dagger}$University of Amsterdam \\
  $^{\ast}$Independent Researcher \\
  \texttt{kumiko.nakajima@student.uva.nl,
  janzuiderveld@gmail.com,
  s.pezzelle@uva.nl}
}

\begin{document}
\maketitle
\begin{abstract}
Large language models (LLMs) are increasingly used in verbal creative tasks. However, previous assessments of the creative capabilities of LLMs remain weakly grounded in human creativity theory and are thus hard to interpret.
The widely used 
Divergent Association Task (DAT) focuses on \textit{novelty}, ignoring \textit{appropriateness}, a core component of creativity. We evaluate a range of state-of-the-art LLMs on DAT and show that their scores on the task are lower than those of two baselines that do not possess any creative abilities, undermining its validity for model evaluation.
Grounded in human creativity theory, which defines creativity as the combination of novelty and appropriateness, we introduce Conditional Divergent Association Task (CDAT). CDAT evaluates novelty conditional on contextual appropriateness,
separating noise from creativity better than DAT, while remaining simple and objective. Under CDAT, smaller model families often show the most creativity, whereas advanced families favor appropriateness at lower novelty. We hypothesize that training and alignment likely shift models along this frontier, making outputs more appropriate but less creative. We release the dataset\footnote{\url{https://osf.io/4ng9f/overview?view_only=cbb334c7199a4dbbb431d1a75db1ed2f}} and code\footnote{\url{https://github.com/knakajima1225/beyond_divergent_creativity}}. 
\end{abstract}

\section{Introduction}
Large language models (LLMs) are increasingly deployed in contexts that require not only linguistic competence, but also the ability to generate outputs that are novel and contextually meaningful, such as storytelling (e.g.,~ \citealt{mirowski2022cowritingscreenplaystheatrescripts};~ \citealt{yuan_wordcraft_2022}), idea generation \citep{Pu_2025}, and writing assistance \citep{chakrabarty_art_2024}.

Despite the growing use of LLMs on these tasks, there is still no consensus on benchmarks and metrics that ground creativity evaluation in both novelty and appropriateness, two key aspects of human creativity~\citep{runco_standard_2012}. Such an assessment is crucial because it grounds creativity in human intuitive judgments and provides a meaningful, interpretable normative target.

To address this gap, we build on lexical Divergent Thinking (DT), a fundamental component of human creative generation. 
DT is a well-established construct in human creativity research that captures the ability to produce multiple novel and varied solutions to open-ended problems \citep{guilford_1967, guilford-1950-creativity}. Divergent Association Task (DAT; \citealt{olson_naming_2021}) is a standard test to measure divergent creative behavior in humans, in which participants produce 10~nouns that are as different from one another as possible; the creativity score is calculated as the average pairwise semantic distance.
While \citet{chen_probing_2023} and \citet{bellemare-pepinDivergentCreativityHumans2026} systematically investigated LLM performance on DAT, this metric, by design, emphasizes novelty (semantic distance) and does not assess whether the produced words are appropriate to any context. As a result, stochastic or random outputs can attain very high DAT scores, rivaling strong LLMs while lacking meaning or usefulness. In short, a novelty-only metric risks conflating randomness with creativity, limiting its validity for LLM assessment. 

To address the limitations of the original Divergent Association Task (DAT) in this context, we propose a simple yet objective Conditional Divergent Association Task (CDAT). CDAT aligns with the widely accepted view in psychology that creativity requires both novelty and appropriateness~\citep{cropley_praise_2006, runco_standard_2012, diedrich_are_2015}. 
Grounded in the prior definitions of creativity based on novelty and appropriateness~\citep{heinen_semantic_2018}, CDAT prompts a model to generate 10~words that are semantically diverse yet relevant to a cue word. Performance is scored on novelty (average pairwise semantic distance among words) and appropriateness (average distance from the cue, penalizing irrelevant words).
We introduce a Pareto Front-based metric to evaluate models' capability to balance novelty and appropriateness and to score their creativity in the context of the given cue.

The paper advances research on machine creativity through the following contributions:
\begin{itemize}
    \item We critically evaluate the validity of Divergent Association Task (DAT). In particular, we show that two non-creative baselines achieve a higher score than the state-of-the-art models on DAT, invalidating the test for LLM creativity evaluation. 
    
    \item We introduce Conditional Divergent Association Task (CDAT) to address DAT's limitations. 
    
    \item We show that CDAT identifies more nuanced, systematic patterns in models' creativity profiles: smaller model families generally show the most creativity, whereas advanced families favor appropriateness at lower novelty.
\end{itemize}

\section{Related Work}
\subsection{Assessing Human Creativity}
To anchor our work in established human creativity constructs, we draw on the foundational framework proposed by \citet{guilford_1967}.
In it, a core component of human creativity is represented by
\textit{Divergent Thinking} (DT), namely, the ability to produce multiple, varied, and novel solutions to open-ended problems. Crucially, DT is quantitatively measurable and excludes unrelated abilities~\citep{guilford_1967, guilford-1950-creativity}, which makes it attractive for any study aiming to obtain quantitative estimates of this fundamental human skill. 
Because of that, measuring DT has become the dominant psychometric approach in the assessment of creativity~\citep{kaufman_essentials_2008}. Examples of DT-based assessments are Alternative Uses Task (AUT; \citealt{guilford_1967}), which tests participants' ability to come up with as many uses as possible for a common object under a time constraint, or Torrance Tests of Creative Thinking (TTCT; \citealt{torrance-1990-ttct-verbal}), which comprise a standardized set of open-ended tasks, including verbal exercises (e.g., imagining the consequences of improbable events) and figural exercises (e.g., completing incomplete drawings), with responses scored by human raters.
Finally, one of the most recent tests, Divergent Association Task (DAT;~\citealt{olson_naming_2021}), measures the semantic distance between words generated by participants: the greater the semantic divergence, the more creative the participants are considered to be. Owing to its strong validity (its scores correlate with established DT tasks such as AUT) and its high test-retest reliability, DAT has become one of the most widely used measures of human creativity. Such semantic-distance based measures, such as DAT scores and related metrics, have been found to human real-world creative behavior (\citealt{georgiev_enhancing_2018}; \citealt{patterson_multilingual_2023}, \citealt{olson_naming_2021}) and can be extended to more contextually richer texts \citep{johnson_divergent_2023}.

While \textit{novelty}, typically linked to divergent thinking, is a crucial component of creativity, it alone is not sufficient. Creative ideas must also be appropriate in a context~\citep{cropley_praise_2006, runco_standard_2012, diedrich_are_2015}. In line with this, \citet{heinen_semantic_2018} operationalize creativity along two key dimensions: \textit{novelty} and \textit{appropriateness} in the verb-generation task \citep{prabhakaran_thin_2014}. In this task, participants produced a verb in response to a cue noun, responses were rated as \textit{appropriate}, reflecting how understandable and accessible the association was, and as \textit{novel}, reflecting how original or unexpected it was (i.e., the verb could be unrelated to the noun).
Random answers scored highest on novelty but lowest on appropriateness, whereas common answers showed the opposite pattern. Crucially, creative answers were those balancing both aspects, lying between random and common responses. 
In this work, we build on this previous research and consider creativity as involving a trade-off between these two aspects.

\subsection{Testing Creativity in LLMs}

Studies have investigated LLM creativity in various domains, such as text and story generation (e.g.,~ \citealt{ismayilzada_evaluating_2024};~\citealt{marco_small_2025}; \citealt{lu_ai_2025}), problem solving (\citealt{chen_deepmath-creative_2025}; \citealt{tian_macgyver_2025}), creative drawing \citep{nath_pencils_2025}, and across domains \citep{cao_evaluating_2025}. A few studies have built on DT-related evaluation, which provide principled, human-relevant measures of creativity. 
For example, \citet{stevenson_putting_2022} adapted AUT to rate the creativity of GPT-3; ~\citet{nath-etal-2024-creative-process} examined how LLMs explore ideation spaces within AUT. 
\citet{zhao_assessing_2025} proposed a TTCT-inspired framework with LLM-based evaluation across prompts and role-play scenarios. Furthermore, \citet{chakrabarty_art_2024} built on TTCT to propose Torrance Test of Creative Writing (TTCW) to evaluate creative writing, and \citet{li_automated_2025} subsequently improved the evaluation by scoring against high-quality reference texts. 
Unlike AUT and TTCT, DAT is scored objectively via semantic distance without human raters and involves no time constraints, making it a practical tool for assessing LLM creativity. Accordingly, \citet{bellemare-pepinDivergentCreativityHumans2026} systematically compared human and LLM DAT performance and found that GPT-4~\citep{openai_gpt-4_2024} and Llama~3.1-8B~Instruct~\citep{meta-llama-3-1} outperformed humans.
\citet{chen_probing_2023} reported results indicating that a random baseline sampling from WordNet~\citep{miller_wordnet_1995} generally scored highest on DAT, followed by GPT-4 and GPT-3.5~Turbo. 

Several benchmarks have incorporated a two-criterion definition of creativity, solving some of the problems of novelty-only assessment. For instance,~\citet{mclaughlin_aidanmclaughlinaidanbench_2025} developed AidanBench, a set of open-ended idea-generation tasks evaluated for both novelty and coherence. Similarly, \citet{lu_benchmarking_2025} assessed LLM creative problem solving in code generation, combining novelty of solutions with correctness and constraint adherence. While promising, these approaches are influenced by many confounding factors, such as word knowledge and integration, and cannot easily be interpreted based on human creativity. Building on \citet{heinen_semantic_2018} and leveraging the objectivity and simplicity of DAT, we propose Conditional Divergent Association Task (CDAT): a human-interpretable test that evaluates both novelty and appropriateness under a controlled condition by grounding model generations on a cue word as a minimal context. CDAT measures a model’s lexical divergent ability to produce divergent yet contextually related words. To our knowledge, it represents the first systematic application of the scoring method that captures this two-fold criterion in a single metric to LLMs.

\section{Is DAT a Valid Measure of Creativity?}
\label{sec:dat}

Building on previous work~\citep{bellemare-pepinDivergentCreativityHumans2026}, we first examine the performance of current LLMs on the standard Divergent Association Task (DAT). Extending this research, we then question whether DAT provides a valid measure of creativity in LLMs, given that, according to human psychology studies, it primarily captures only the \textit{novelty} dimension of creative ability. We achieve this by extensively testing a range of state-of-the-art (SotA) LLMs of various types and sizes, exploring the role of temperature in sampling words. We compare the results of our models against two baselines, one at the opposite end of the spectrum: a fully \textit{task-agnostic} baseline that samples random words from WordNet; and a fully \textit{task-aware} baseline that `knows' the task and how to achieve a high score. While they use different strategies, both of these baselines are \textit{non-creative}, being based on random or `cheating' behaviors, respectively.

\subsection{Task}
\paragraph{Specification.}
In Divergent Association Task (DAT), participants are asked to generate a set of unrelated words, and creativity is quantified by calculating the average semantic distance between them using computational models of word meaning. In particular, the greater the semantic divergence, the higher the DAT score they achieve and, in turn, the more creative they are considered to be. Here, we prompt models to either generate or sample 10 words that are as unrelated as possible. We then evaluate how divergent these words are by computing their embedding dissimilarity (see §~\ref{subsec:evaluation}).

\paragraph{Prompt.}
We use the prompt from \citet{bellemare-pepinDivergentCreativityHumans2026}, with minimal formatting and grammatical constraints to ensure that outputs are \emph{singular common nouns}, listed one per line: \textit{``Please enter 10 words that are as different from each other as possible, in all meanings and uses of the words.''} Full prompt text and formatting rules are provided in Appendix~\ref{subsec:dat-prompt}.

\subsection{Evaluation}
\label{subsec:evaluation}

\paragraph{Validity filtering.}
Following the original DAT procedure~\citep{olson_naming_2021}, we consider the \emph{first seven} valid items in each model response. An item is valid if it is a properly spelled English word, \emph{single-word common noun} (no proper nouns, no multi-word expressions, no numerals). We implement lemmatization and POS-based noun filtering with NLTK~\citep{bird_natural_2009}, accepting a word only if it is tagged as a noun or associated with at least one noun synset in WordNet. Exact rules and exceptions are detailed in Appendix~\ref{subsec:dat-eval}.

\paragraph{Scoring.}
For each model response, we compute novelty over the \emph{first seven valid words} only. Let $\mathbf{e}_{w_i}$ and $\mathbf{e}_{w_j}$ be embedding representations of two generated nouns $w_i$ and $w_j$. The per-pair novelty is
\[
\mathrm{Nov}(w_i,w_j)=100\cdot\bigl(1-\cos(\mathbf{e}_{w_i},\mathbf{e}_{w_j})\bigr)\in[0,200],
\]
and the per-response novelty is the average over all unordered pairs among the first seven valid words; higher values indicate greater semantic dissimilarity between the words and therefore higher novelty. If fewer than seven valid words remain for a response, that (model, response) instance is dropped.

\begin{figure*}[t!]
\centering
\includegraphics[width=1\textwidth, trim=20 0 0 0, clip]{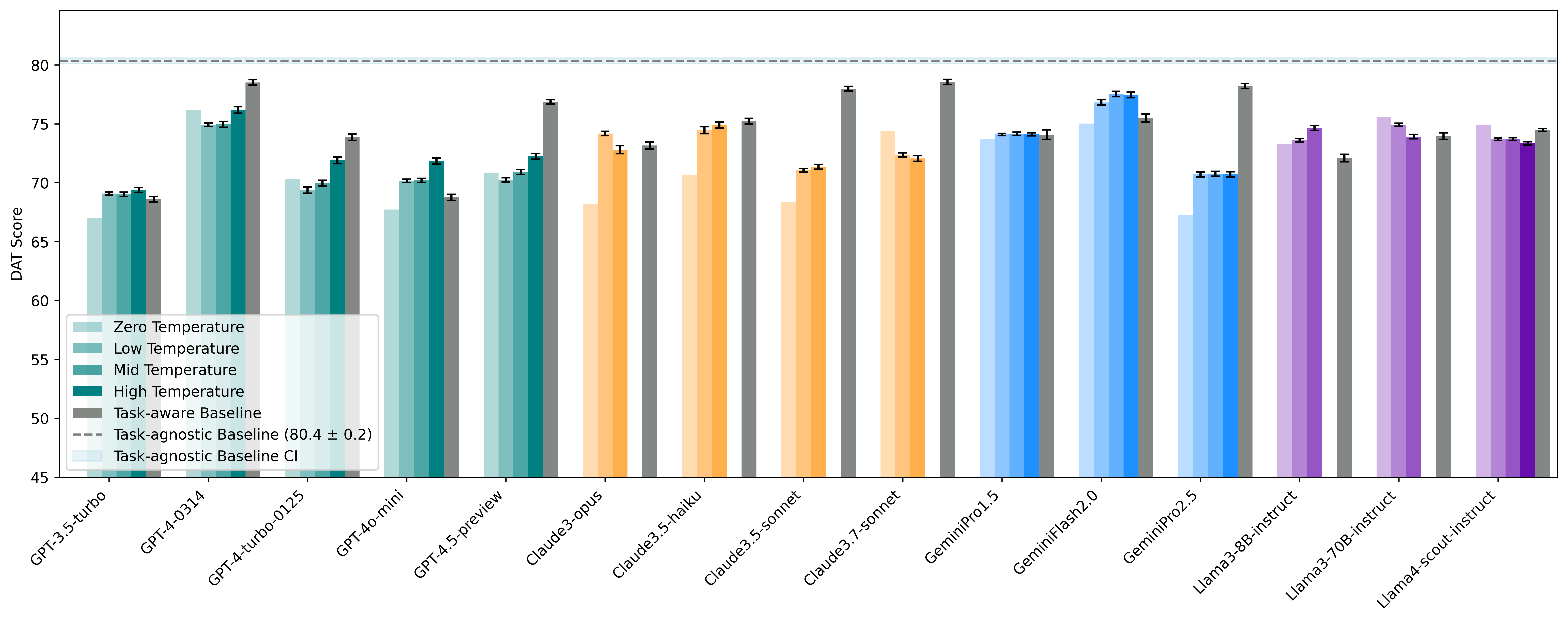}
\caption{\textbf{Mean DAT scores across temperature settings with the DAT prompt}, compared with two non-creative baselines: \textit{task-agnostic} (horizontal dotted line) and \textit{task-aware} (gray bars). Error bars denote 95\% CIs for models; the shaded band shows the 95\% CI for the \emph{task-agnostic} baseline. All valid responses are included. Temperature settings: Low ($t{=}0.5$), Mid ($t{=}1.0$), High ($t{=}1.5$; not supported by all models). The \emph{task-aware} baseline is evaluated at Mid only ($t{=}1.0$).}
\label{fig:random-baseline}
\end{figure*}

\paragraph{SBERT vs.\ GloVe.}
DAT, in its standard implementation, uses GloVe \citep{pennington_glove_2014} to obtain word representations. However, GloVe is sensitive to frequency artifacts: static embeddings trained on global co-occurrence statistics tend to encode word frequency (e.g., vector norms and geometry correlate with frequency), which is especially problematic for rare words, and can inflate novelty for lists containing uncommon nouns~\citep{schnabel_evaluation_2015, chen_probing_2023}. In contrast, we propose using a BERT-based model for embedding words. BERT's subword tokenization and contextual pretraining provide more stable lexical representations for infrequent items (\citealt{devlin_bert_2019}; \citealt{sennrich_neural_2016}). Sentence BERT (SBERT) is a variant of BERT that efficiently creates single, meaningful vector embeddings for entire sentences, making it superior for semantic similarity tasks \citep{reimers_sentence-bert_2019}. Here, we assume that SBERT provides an embedding space that still yields meaningful representations for single words. Thus, from here on, we report the results obtained with SBERT-based embeddings.

\paragraph{Baselines.}
We include two non-creative baselines, both evaluated under the same filtering and scoring pipeline detailed above:
\begin{enumerate}[leftmargin=*]
    \item \textbf{Task-agnostic baseline.} We uniformly sample 10 nouns from WordNet (excluding multi-word phrases, proper nouns, and numerals) and compute DAT scores as above. We repeat this procedure 500 times to obtain a sampling distribution.
    \item \textbf{Task-aware baseline.} We instruct the same models used in §\ref{subsec:models} to explicitly \emph{maximize} the DAT score without any information about the maximum semantic distance constraint: \textit{``Please enter 10 words that maximize the Divergent Association Task (DAT) score. Make a list of these 10 words, a single word in each entry of the list.''} This baseline is evaluated at mid temperature only (\(t{=}1.0\)), with 500 generations per model. By design, this condition is \emph{non-creative}: it optimizes the test objective directly rather than producing novel content under task-general instructions.
\end{enumerate}

\subsection{Models}
\label{subsec:models}

We evaluate proprietary and open-source SotA LLMs. Concrete provider/model identifiers and versioned API endpoints are documented in Appendix~\ref{sec:selected-models}. For each model, we generate 500 responses at three temperatures: low (\(t{=}0.5\)), mid (\(t{=}1.0\)), and high (\(t{=}1.5\)) plus a single deterministic sample at \(t{=}0.0\). Some model families do not support \(t{=}1.5\) or fail to produce reliably parseable outputs at that setting; those exceptions and all other implementation details (prompt formatting, alternative sampling strategies) are provided in Appendix~\ref{subsec:dat-models}.

\subsection{Results}
\label{subsec:dat-results}

Unless otherwise specified, all pairwise comparisons are conducted using two-sided independent-sample t-tests, with multiple comparisons corrected via Benjamini--Hochberg false discovery rate (FDR) procedure \citep{benjamini1995}. Outliers exceeding three standard deviations from the mean are removed from each model’s score distribution, following~\citet{bellemare-pepinDivergentCreativityHumans2026}. Here, we report statistical tests for between-condition comparisons. 

\paragraph{Replication with GloVe.}
Using GloVe, we reproduce the qualitative ordering reported by \citet{bellemare-pepinDivergentCreativityHumans2026} on the overlapping model subset. Pairwise contrasts and temperature effects match the original patterns; full estimates and tests appear in our GitHub repository. 

\paragraph{Switching to SBERT.}
By regressing the DAT score on word surprisal, as in \citet{chen_probing_2023}, we observe that SBERT-based DAT scores are less sensitive to word frequency than GloVe-based scores, although the two measures remain strongly aligned (Spearman’s $\rho{=}0.73$). This supports our hypothesis that SBERT provides a more robust yet broadly equivalent replacement for GloVe, which was originally designed for single-word embeddings; details are in Appendix~\ref{subsec:dat-sbert}. We therefore experiment with SBERT as our main embedding model.

\paragraph{Temperature.}
Across low, mid, and high settings, temperature effects are mixed and family-specific: some families show higher \emph{novelty} with temperature while others decrease, with no consistent trend across families. We report tests for all available temperatures per model in our GitHub repository. 

\paragraph{Baselines.}
The central finding is that, under SBERT scoring, no family exceeds the task-agnostic baseline at any tested temperature (Figure~\ref{fig:random-baseline}). At mid temperature, within-family contrasts show the task-aware baseline yields higher mean \emph{novelty} than the corresponding DAT condition for most families, with significant gaps in the majority of cases (full data in our GitHub repository). These results indicate that DAT is sensitive to random sampling and prompt familiarity rather than isolating creative ability. This motivates a contextualized evaluation (§~\ref{sec:cdat}).

\section{Conditional Divergent Association Task}
\label{sec:cdat}

To address the DAT limitations observed in §~\ref{sec:dat}, namely that high divergence can be achieved while being task-agnostic, we introduce \textit{Conditional Divergent Association Task} (CDAT). CDAT places generation under an explicit contextual constraint: outputs must remain semantically associated with a given cue while being mutually diverse. This allows us to quantify creativity while preserving a principled link to context, aligned with the definition of \citet{heinen_semantic_2018}. Following \citet{diedrich_are_2015}, we place slightly greater emphasis on novelty and evaluate it conditional on appropriateness, reflecting the view that appropriateness serves as a second-order criterion of creativity, particularly for novel responses.

In this section, we (i) formalize the task (§~\ref{sec:cdat-task}), (ii) define the evaluation protocol, including a minimal appropriateness criterion and the CDAT score (§~\ref{sec:cdat-eval}), (iii) describe the models and sampling setup (§~\ref{sec:cdat-models}), and (iv) present results, starting from the 2D appropriateness–novelty landscape and its Pareto front, then reporting scalar CDAT scores (§~\ref{sec:cdat-results}); complementary diagnostics (Elbow distance and Distance to Human) are summarized briefly and analyzed in depth in Appendix~\ref{subsec:cdat-appdx-reslt-diagnostics}. Throughout, we adopt an operational view of creativity as producing \emph{diverse yet contextually appropriate} outputs; the formal details of how we quantify this view follow below. We emphasize that this operationalization is a pragmatic choice for \emph{quantifiability}, not a claim to be a complete account of creativity; the 2D analyses help reveal trade-offs that future work may refine.

\subsection{Task} \label{sec:cdat-task}

\paragraph{Specification.}
Given a cue \textit{noun}, the model must generate 10 mutually dissimilar nouns that are each semantically associated with the cue. We evaluate only single-word common nouns (as in §~\ref{sec:dat}), applying the same validity rules (duplicates, non-English words, multiword phrases, proper nouns, etc.).

\paragraph{Prompt.}
We elicit outputs with the following instruction (full prompt in Appendix~\ref{subsec:cdat-appdx-task}):
\emph{``Please enter 10 words that are as different from each other as possible, in all meanings and uses of the words, yet semantically associated with the following cue word: \{cue\}.''} The rest of the prompt is the same as DAT.

\begin{table}[t!]
\centering
\begin{tabular}{p{7.5cm}}
\toprule
\textbf{Model:} Gemini 2.0 Flash \\
\textbf{Temperature:} 1.0 \\
\textbf{Cue:} \textit{unity} \\
\midrule
\textbf{Response:} \\
1. fragmentation, 2. diversity, 3. harmony, \\
4. whole, 5. difference, 6. separation, \\
7. aggregate, 8. coalition, 9. relationship, \\
10. inclusion \\
\bottomrule
\end{tabular}
\caption{Example CDAT output by Gemini 2.0 Flash.}
\label{tab:cdat-example}
\end{table}

\paragraph{Cues.}
We derive cue nouns from the 10{,}000 most frequent words in the Brown corpus~\citep{francis1979brown} by lemmatizing to singular nouns via WordNet, validating common-noun status with WordNet synsets and NLTK POS tags, and excluding multiword items, pure digits, and single characters. Remaining candidates are filtered with GPT-4.1~nano to remove verbs and proper nouns. This yields 539 unique cues (mean frequency ${=}28.94$, mean length ${=}7.2$). See Appendix~\ref{subsec:cdat-appdx-task} for more details.

\subsection{Evaluation} \label{sec:cdat-eval}
\paragraph{Novelty.}
Our novelty metric is \emph{identical} to DAT's novelty metric in §~\ref{sec:dat}: for each (model, cue) list of valid outputs, we compute the standard DAT score, then average across cues to obtain mean novelty.

\paragraph{Appropriateness.}
For each (model, cue), we compute appropriateness over the first seven valid words only. Let $\mathbf{e}_c$ and $\mathbf{e}_w$ be word representations for the cue and a generated noun $w$. The per-word appropriateness is
\[
\mathrm{App}(c,w)=100\cdot\bigl(1+\cos(\mathbf{e}_c,\mathbf{e}_w)\bigr)\in[0,200],
\]
and the per-cue appropriateness is the average over the first seven valid words; higher values indicate greater cue–word semantic \textit{similarity} and therefore higher appropriateness. If fewer than seven valid words remain for a cue, that (model, cue) instance is dropped.

\paragraph{Baselines.}
We use two task-agnostic baselines: Random and Common. The \textit{Random} baseline draws 10 nouns per cue \emph{independently of the cue} from the same evaluation vocabulary as §~\ref{subsec:evaluation}; we generate 500 such cue lists. The \textit{Common} baseline prompts GPT-4.1~nano with: \emph{``enter 10 words that are most semantically associated with the following cue word: \{cue\}''} for each CDAT cue, yielding a purely association-driven set. The Random baseline is used to define the minimal appropriateness criterion (gate) and as a reference in 2D plots; the Common baseline is used only for \emph{diagnostic} 2D analyses (Pareto front, Elbow distance).

\paragraph{Appropriateness gate (minimal criterion).}
We use a gating mechanism to judge if the models generate words that are considered appropriate or semantically related to the cue.
For each \textit{model}, we compare its distribution of per-cue appropriateness values to the task-agnostic Random baseline using two-sided Welch's $t$-tests and control FDR at $\alpha{=}0.001$ within each temperature. A model is retained only if (i) its FDR-adjusted $p$-value is below $0.001$ and (ii) its mean appropriateness exceeds that of the Random baseline, $\overline{\mathrm{App}}_{\mathrm{model}}{>} \overline{\mathrm{App}}_{\mathrm{random}}$. All models remain significant at more lenient thresholds $\alpha\in[0.05, 0.01, 0.005]$, so we report results using $\alpha{=}0.001$, the strictest level usually used in empirical studies.
This criterion is \emph{not} itself the creativity score; it ensures that novelty is interpreted under a minimal standard of contextual alignment. 

\paragraph{Scoring.}
Our primary scalar measure of creativity is the mean novelty (DAT-style) for each \textit{model}, conditional on passing the appropriateness gate. Models that fail the gate are flagged and excluded from the CDAT ranking. In our experiments, all models `pass' the gate for each tested temperature. 

\paragraph{2D trade-off visualization and Pareto front.}
Because creativity involves balancing novelty with contextual appropriateness, we also analyze the 2D plane of mean appropriateness vs.~mean novelty for each model. We overlay the Pareto front of non-dominated model points to highlight optimal trade-offs. The task-agnostic, Random and Common baselines serve as anchors (maximal novelty vs.\ maximal appropriateness, respectively). These 2D views are \emph{descriptive} complements to the primary scalar CDAT score.

\paragraph{Diagnostic metrics (non-primary).}
We compute two additional diagnostics: (i) Elbow distance, the average signed Euclidean distance from each model’s individual points to the line connecting the two extreme task-agnostic baselines (Random and Common) means; and (ii) Distance to Human, the Euclidean distance from a model’s individual point to the aggregated human reference point, obtained by asking 11 annotators to provide words to, on average, 6.4 cues ($n{=}99$; cues{=}70).
These diagnostics aid interpretation and are not used for the primary ranking (more details on the definitions and human data collection in Appendix~\ref{subsec:cdat-appedx-diagnostic}).

\subsection{Models} \label{sec:cdat-models}

\begin{figure}[t!]
\includegraphics[width=1\columnwidth, trim=0 0 0 20, clip]{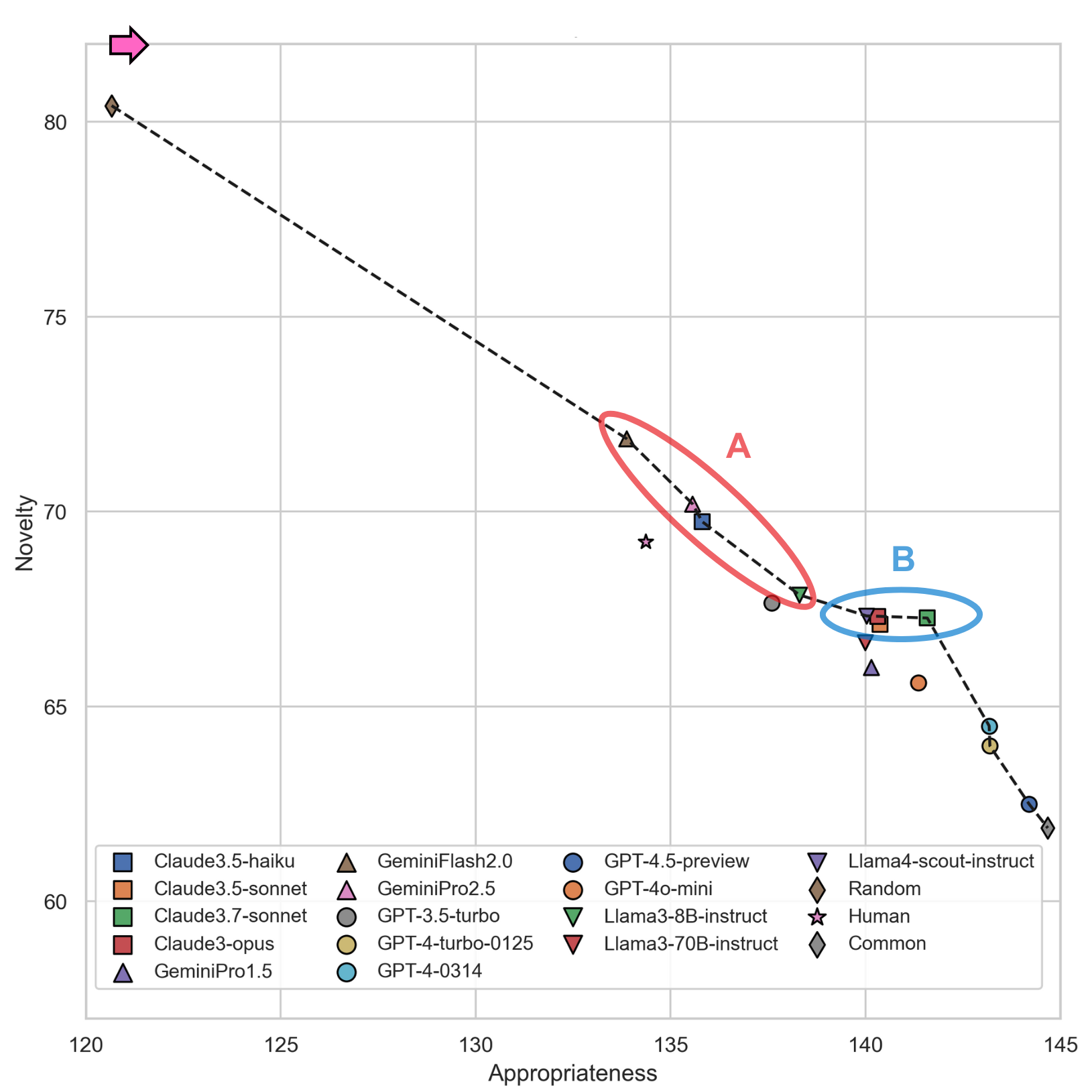}
\caption{\textbf{CDAT appropriateness–novelty landscape (mid temperature).} Solid markers show model means. The pink arrow indicates the ideal region under our operationalization (novelty evaluated after passing the appropriateness gate).
The Pareto front (non-dominated models) traces optimal trade-offs. Circle A highlights basic models with high novelty and least appropriateness; Circle B marks advanced models with a balanced novelty-appropriateness trade-off.}
\label{fig:pareto-mid}
\end{figure}

We evaluate the same set of LLMs as in §\ref{subsec:models} across three temperatures (low${=}0.5$, mid${=}1.0$, high${=}1.5$) using the CDAT prompt above (noun-only constraint enforced by our validity rules). For each model, we sample one 10-word list per cue; all models respond to all cues at least once. Model IDs, sampling seeds, and implementation details are the same as DAT (see Appendices~\ref{sec:selected-models} and \ref{subsec:dat-models}). 

\subsection{Results} \label{sec:cdat-results}
\begin{figure}[t!]
\centering
\includegraphics[width=1\columnwidth, trim=0 10 0 100, clip]{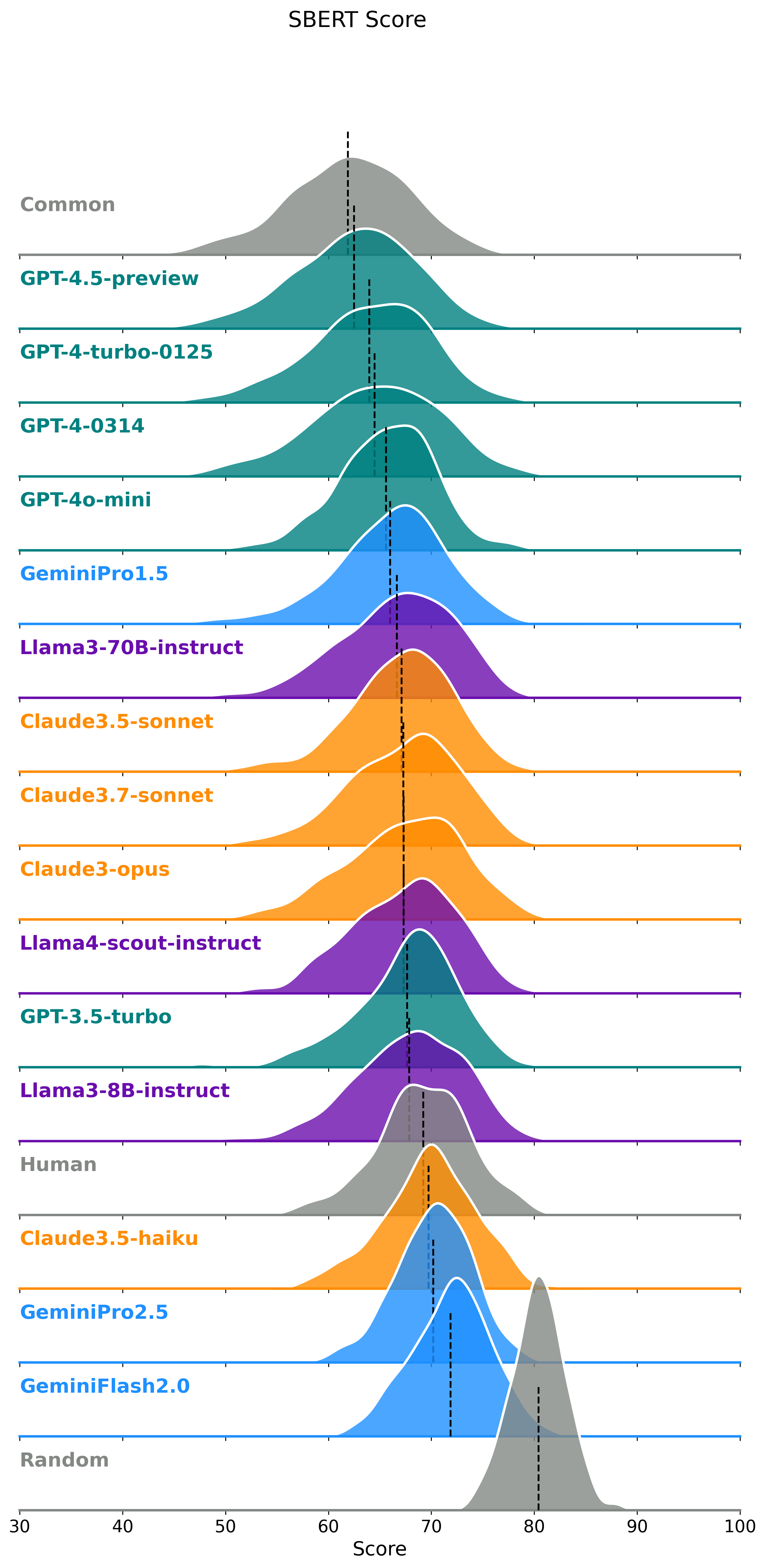}
\caption{\textbf{CDAT \textit{novelty} distributions (mid temperature).} Ridge plot of model novelty (DAT-style) under CDAT. Black vertical lines denote means. Detailed statistics and all temperatures are 
in Appendix~\ref{subsec:cdat-appdx-novelty}.}
\label{fig:cdat-divergence}
\end{figure}

\paragraph{Overview.}
We report mid-temperature results as primary and provide per-temperature tables in Appendix~\ref{sec:cdat-appdx-result}. Figures~\ref{fig:pareto-mid} and~\ref{fig:cdat-divergence} summarize the 2D appropriateness–novelty landscape and the scalar CDAT distributions, respectively. Preliminary counts and sensitivity data are in Appendix \ref{subsec:cdat-appdx-counts}. 

\paragraph{2D landscape and Pareto front.}
Figure~\ref{fig:pareto-mid} shows the mid-temperature 2D landscape. The Pareto front delineates models that are not simultaneously improvable in both dimensions. The pink arrow marks the \textit{ideal region} under our operational view: novelty evaluated after passing the appropriateness gate. Broadly, models occupy a continuum between higher-novelty/lower-appropriateness and more-balanced points with stronger contextual alignment; these cluster-level tendencies are consistent across temperatures (see Appendix~\ref{subsec:cdat-appdx-pareto-temp}). 

\paragraph{Main findings.}
Mid-temperature distributions of scalar CDAT novelty scores are shown in Figure~\ref{fig:cdat-divergence}; full per-temperature significance tests and sensitivity analysis across embedding models appear in Appendices~\ref{subsec:cdat-appdx-novelty} and \ref{subsec:cdat-embedding-power}. We summarize two robust findings.

\textit{(1) Conditioning works across the board.} All models pass the appropriateness gate (two-sided Welch's $t$-test vs.\ Random, $p{<}0.001$ after FDR correction, with higher mean appropriateness than the Random baseline), validating novelty as a creativity signal \emph{conditional} on contextual appropriateness. We confirm that this holds at all temperatures.

\textit{(2) Efficiency-oriented ``basic'' families (smaller, latency-optimized) generally achieve higher CDAT scores; CDAT identifies more nuanced patterns in models' creativity profiles.} ``Advanced'' families (larger, instruction-/reasoning-optimized) typically trade some novelty for stronger appropriateness, yielding more conservative CDAT scores. Both strategies appear on the non-dominated Pareto frontier, and no family simultaneously maximizes both axes; the same pattern holds across temperatures (see Appendix~\ref{subsec:cdat-appdx-pareto-temp}). For reference, basic families score closer to the human reference on novelty when conditioned on contextual appropriateness, while many advanced families adhere to context more strictly than humans typically deem appropriate (where human raters show higher agreement; see Appendix D.2).

\paragraph{Diagnostics (non-primary).}
\textit{Elbow distance.} Relative to the line between Random and Common baselines, results are mixed overall, with a slight edge for advanced families. Rankings across temperatures are provided in Appendix~\ref{subsec:cdat-appdx-reslt-diagnostics}. \newline
\textit{Distance to Human.} Using a human reference ($n{=}99$ over 70 cues), model family distances are mixed overall, but the metric assigns slightly higher values to more basic model families. Given the limited sample and exploratory nature, we treat this comparison as a purely qualitative one (see Appendix~\ref{subsec:cdat-appdx-reslt-diagnostics} for further details).

\section{Discussion and Future Directions}
\paragraph{What CDAT establishes.}
Our results show that a novelty-only assessment such as DAT is vulnerable to both random sampling and prior familiarity with the objective, which undermines its validity for LLM creativity evaluation (§~\ref{sec:dat}; Figure~\ref{fig:random-baseline}). 
More broadly, our findings indicate that naive adaptations of human creativity tests such as DAT, as used in prior work (\citealt{bellemare-pepinDivergentCreativityHumans2026}; \citealt{chen_probing_2023}), pose substantial limitations.
CDAT addresses this by conditioning novelty on a minimal appropriateness criterion (§4.2), thereby 
separating noise from creativity more effectively while preserving objectivity and simplicity.

\paragraph{Interpreting the trade-off.}
Under this operationalization, we observe systematic, family-level profiles in the appropriateness–novelty plane (Figure~\ref{fig:pareto-mid}) and corresponding scalar CDAT distributions (Figure~\ref{fig:cdat-divergence}), consistent across temperatures (Appendix~\ref{sec:cdat-appdx-result}). 
Across families, we see a stable pattern: smaller models tend to occupy higher-novelty regions with reduced appropriateness, while larger models shift toward higher appropriateness with lower novelty (Figure~\ref{fig:pareto-mid}). 
We hypothesize that scaling, instruction tuning, and safety alignment alter sampling preferences along the appropriateness–novelty frontier, prioritizing conservative, predictable outputs. 
However, this remains a correlational interpretation: targeted ablations on, for example, alignment objectives and decoding policies are needed to establish causal mechanisms. Discussion on prompting and decoding, and proposed mechanism tests as capability probes, alongside temperature-based diagnostics of novelty constraints, are in Appendix~\ref{subsec:cdat-appdx-ablation}.

\paragraph{Scalar scores vs.\ 2D views.}
We retain the scalar CDAT score for comparability and ranking: the mean DAT-style novelty, conditional on passing the appropriateness gate. The 2D landscape and its non-dominated set are descriptive complements that make trade-offs visible and help diagnose why similarly ranked systems produce qualitatively different behaviors. In practice, we recommend reporting: (i) gate pass rates, (ii) scalar CDAT with confidence intervals, and (iii) the model mean in the 2D plane anchored by task-agnostic baselines. Diagnostics like Elbow distance or Distance-to-Human can be used to gain qualitative insights.

\paragraph{Measurement choices and their implications.}
Two design decisions matter. First, the appropriateness gate uses the strictest, model-specific comparison against Random with a directional constraint (Welch’s $t$, $\alpha{=}0.001$) to ensure that novelty is interpreted only under minimal contextual alignment and to reduce residual gaming risk. Post-hoc power analyses appear in Appendix~\ref{subsec:cdat-embedding-power}.
How to define a minimal level of appropriateness indeed remains an open question; future work could, for example, flag responses outside the Random baseline confidence interval for targeted human review or LLM-based semantic checks.
Second, we adopt SBERT-based lexical representations to reduce frequency artifacts known to affect GloVe; however, representation choice is itself an experimental factor. Therefore, we conduct sensitivity analysis across SBERT, GloVe, fastText \citep{bojanowski_enriching_2017}, and BERT \citep{devlin_bert_2019} in  Appendix~\ref{subsec:cdat-embedding-power}.

\paragraph{Human relevance and scope.}
Just as with DAT, whether a valid and reliable test for humans is valid for LLMs is an open question: recent evidence suggests that any general factor in LLM performance may emerge primarily on tasks that rely on crystallized, text-based knowledge rather than mirroring human-like inter-task correlations \citep{ilic_evidence_2024}. We emphasize that CDAT evaluates behavioral outputs, not cognition, and targets a minimal component of divergent verbal creativity. It is best viewed as a measure of crystallized, verbal creativity that should generalize primarily to other verbal divergent-thinking tasks (e.g., AUT and verbal TTCT) where semantic knowledge is central, and where LLMs are most often deployed. At the same time, CDAT is not a universal predictor of creativity and should be seen as complementary to context-rich creativity assessments that additionally involve domain knowledge, planning, or any kind of long-context understanding; related discussion and future scope extensions are given in Appendix~\ref{subsec:cdat-appdx-scope}.

\section{Conclusions}
In this paper, we focused on an important and timely question: \textit{How can we genuinely assess LLMs' creative abilities?}
We showed that the standard DAT assessment, being focused only on the novelty component of creativity, is 
insufficient for LLM assessment.
Indeed, DAT turned out to be sensitive to task-agnostic random generation and familiarity with the task objective and failed to genuinely capture theory-driven, human-interpretable creativity.

Building on these findings, we introduced
Conditional Divergent Association Task (CDAT), a two-component measure of creativity
which operationalizes creativity as \textit{novelty} conditional on \textit{appropriateness} to a minimal context. CDAT reinstates contextual fit as a first-class requirement while retaining the objectivity and simplicity of semantic-distance-based evaluation. 
We showed that CDAT separates noise from creativity more effectively than DAT.
Evidence comes from the appropriateness gate and the 2D novelty–appropriateness landscape with Pareto structure.
Using CDAT, we find systematic creativity profiles across model families. Simpler families tend toward higher novelty with reduced appropriateness. Larger and more aligned families occupy operating points with higher appropriateness and lower novelty. Under our operational definition, such models can appear less creative because scaling, instruction tuning, and safety alignment likely shift them along the novelty–appropriateness frontier toward safer, more predictable outputs, although this is a hypothesis, not a causal proof.

Overall, CDAT offers a pragmatic, theory-aligned step toward measuring creativity in LLMs. The observed family-wise trade-offs provide a map rather than a verdict. Future work should turn this map into levers through ablations, calibrated prompting, and cross-task validation to separate creative capability from alignment preferences.

\section{Limitations}
\paragraph{Human sample size.}
The human average is based on a limited sample size ($n{=} 99$), which may affect the stability of human performance estimates and, in turn, our analysis. 
\paragraph{Cue selection.}
While we aim for lexical diversity, the cue word selection process may have introduced unintended variability in model scores. 
\paragraph{Model and task setups.} 
Our evaluation is restricted to a specific task formulation, prompt, and set of models. Results may not extend to other architectures, alternative prompting strategies, different decoding methods, or different measures to quantify semantic relatedness. 
\paragraph{Creativity in applied contexts.} 
CDAT captures only one component of creativity: lexical divergent thinking under minimal contextual constraints. Creativity in applied settings often involves richer context, longer outputs, multimodal grounding, or other domain-specific knowledge, which are not represented here.
\paragraph{Open challenge.} 
Additionally, although we aim for human-interpretable, theory-driven diagnostics, creativity remains a multifaceted construct, and alternative metrics or human-centered evaluations may emphasize different aspects of creativity. Defining measures that capture creative performance across domains and tasks remains an open challenge. 
Despite these limitations, CDAT provides a human-inspired framework for assessing creativity in LLMs by jointly considering novelty and appropriateness. We hope this work contributes to more systematic comparisons of creative performance and motivates further development of metrics and methods for evaluating creativity in artificial systems.

\section*{Acknowledgments}

We thank the members of the Dialogue Modelling Group (DMG) from the University of Amsterdam and Claire Stevenson for the helpful feedback provided at different stages of this project.

We acknowledge the use of generative AI tools (ChatGPT\footnote{\url{https://chatgpt.com/}, accessed 23 January 2026}, Cursor\footnote{\url{https://cursor.com/home?from=agents}, accessed 23 January 2026}) for assistance in manuscript refinement, code structuring, and debugging.

\bibliography{reference_clean}

\appendix
\clearpage

\begin{table*}[!t]
\centering
\small
\setlength{\tabcolsep}{5pt}
\renewcommand{\arraystretch}{1.1}
\begin{tabular}{|l|l|l|c|c|}
\hline
\textbf{Model name} & \textbf{Source} & \textbf{Model ID} & \textbf{Year} & \textbf{Distinction} \\
\hline
GPT-3.5-turbo & OpenAI & gpt-3.5-turbo & 2022 & Middle \\
GPT-4 & OpenAI & gpt-4-0314 & 2023 & Advanced \\
GPT-4-turbo & OpenAI & gpt-4-0125-preview & 2024 & Advanced \\
GPT-4o-mini & OpenAI & gpt-4o-mini & 2024 & Basic \\
GPT-4.5 & OpenAI & gpt-4.5-preview-2025-02-27 & 2025 & Advanced \\
Claude 3 Opus & Anthropic & claude-3-opus-20240229 & 2024 & Advanced \\
Claude 3.5 Haiku & Anthropic & claude-3-5-haiku-20241022 & 2024 & Basic \\
Claude 3.5 Sonnet & Anthropic & claude-3-5-sonnet-20241022 & 2024 & Middle \\
Claude 3.7 Sonnet & Anthropic & claude-3-7-sonnet-20250219 & 2025 & Middle \\
Gemini 1.5 Pro & Google & gemini-1.5-pro & 2024 & Middle \\
Gemini 2.5 Pro & Google & gemini-2.5-pro-preview-03-25 & 2025 & Advanced \\
Gemini 2.0 Flash & Google & gemini-2.0-flash & 2025 & Basic \\
Llama 3.1-8B Instruct & Meta & meta-llama/Meta-Llama-3.1-8B-Instruct-Turbo & 2024 & Basic \\
Llama 3.1-70B Instruct & Meta & meta-llama/Meta-Llama-3.1-70B-Instruct-Turbo & 2024 & Middle \\
Llama 4 Scout Instruct & Meta & meta-llama/Llama-4-Scout-17B-16E-Instruct & 2025 & Advanced \\
\hline
\end{tabular}
\captionsetup{justification=centering}
\caption{Model information.}
\label{tab:model-info}
\end{table*}

\section{Selected Models}
\label{sec:selected-models}

Table~\ref{tab:model-info} lists each model’s name, source, model ID, year, and its
classification as basic (smaller, latency-optimized), middle, or advanced (larger,
instruction-/reasoning-optimized), following \citet{payne-alloui-cros-2025-strategic}.
\section{DAT Method}
\label{sec:dat-method}

\subsection{Prompt}
\label{subsec:dat-prompt}

We use the following prompt for DAT:

\emph{``Please enter 10 words that are as different from each other as possible, in all meanings and uses of the words. Rules: Only single words in English. Only nouns (e.g., things, objects, concepts). No proper nouns (e.g., no specific people or places). No specialized vocabulary (e.g., no technical terms). Think of the words on your own (e.g., do not just look at objects in your surroundings). Make a list of these 10 words, a single word in each entry of the list.''}

\subsection{Evaluation}
\label{subsec:dat-eval}

As in the original test \citep{olson_naming_2021}, we keep the first seven properly spelled, valid words. In addition to the original setup, we add filtering that cleans each token and keeps it only if NLTK \citep{bird_natural_2009} identifies it as a noun, either by assigning it a noun POS tag (NN/NNS) or by finding at least one noun synset for it in WordNet.
The DAT score is calculated as the average pairwise semantic distance across the seven words, ranging from 0 to 2. The semantic distance is computed from the cosine similarity between embedding vectors, using either GloVe or SBERT. The score is multiplied by 100, which gives the final score ranging between 0 and 200 (the practical range is usually 60–110). 

\subsection{Models}
\label{subsec:dat-models}

Proprietary models are accessed via their official APIs: GPT\footnote{\url{https://platform.openai.com/docs/overview}, accessed 23 January 2026}, Claude\footnote{\url{https://docs.anthropic.com/en/home}, accessed 23 January 2026}, and Gemini\footnote{\url{https://ai.google.dev/}, accessed 23 January 2026}. Llama~Instruct models are accessed through Together AI\footnote{\url{https://www.together.ai/}, accessed 23 January 2026}. We build on the code by \citet{bellemare-pepinDivergentCreativityHumans2026} for this experiment. 

Some model families are excluded from the high-temperature setting ($t = 1.5$): Claude models, which are limited to a maximum temperature of 1.0, and Llama~3.1 models, which do not produce reliably extractable answers at the high temperature. Other sampling strategies, such as top-$k$ and top-$p$, are set to their API defaults (GPT models: top-$p{=}1.0$; Claude: top-$p{=}0.99$; Gemini: top-$p{=}0.95$; Llama: top-$p{=}0.7$, top-$k{=}50$). As in the previous study, we open new conversations or ``chat sessions'' in each iteration to collect samples. 

\subsection{Switching to SBERT}
\label{subsec:dat-sbert}

We compare DAT scores across the selected models using our newly adopted SBERT-based distance metric. To assess potential word-frequency artifacts, we quantify word frequency via surprisal, following \citet{chen_probing_2023}. It is computed for each word as $-\log p(w)$, where $p(w)$ is the add-one–smoothed relative frequency of $w$ in the Brown corpus \citep{francis1979brown}; item-level surprisal is then the mean over its valid words. 

In regressions controlling for model and temperature, the slope of surprisal is consistently smaller for the SBERT-based score than for the GloVe-based score at all temperatures, with all interaction terms between surprisal and SBERT-based metric (vs.\ GloVe) negative ($-1.33$ at low; $-1.51$ at mid; $-1.65$ at high) and highly significant ($p{<}0.001$), indicating that SBERT is substantially less sensitive to word frequency. Nevertheless, SBERT- and GloVe-based DAT score rankings are strongly aligned (Spearman’s $\rho{=}0.73$, $p{=}1.44 \times 10^{-3}$, $\alpha{=}0.01$), suggesting that SBERT provides a more robust yet broadly equivalent replacement for GloVe, which was originally designed for single-word embeddings. The full DAT ranking using SBERT is available in our GitHub repository.

\section{CDAT Method}
\label{sec:cdat-appdx-method}

\subsection{Task}
\label{subsec:cdat-appdx-task}

We use the same models and sampling seeds as in the DAT setup.

\paragraph{CDAT prompt.} 
\emph{``Please enter 10 words that are as different from each other as possible, in all meanings and uses of the words, yet semantically associated with the following cue word: \{cue\}. Rules: Only single words in English. Only nouns (e.g., things, objects, concepts). No proper nouns (e.g., no specific people or places). No specialized vocabulary (e.g., no technical terms). Think of the words on your own (e.g., do not just look at objects in your surroundings). Make a list of these 10 words, a single word in each entry of the list.''}

\paragraph{Cues.} 
We extract cues among the 10{,}000 most frequent words in the Brown corpus by lemmatizing words to their singular noun form using WordNet, validating that words are common nouns by checking both WordNet synsets and NLTK POS tags, and excluding multiword phrases, pure digits, or single characters (mean frequency${=}45.38$, mean length${=}7.2$; $n{=}4256$). The words are sent to GPT-4.1~nano\footnote{Model ID: \texttt{gpt-4.1-nano-2025-04-14}} to filter out verbs and proper nouns and obtain the final 550 cues (unique cues $n{=}539$; mean frequency${=}28.94$ and mean length${=}7.2$) using the following instruction:

\emph{``For each word in the list below, determine whether it is a common noun. Make sure to exclude any words that are verbs or proper nouns (such as specific places or personal names). 
For each word, respond in this format: word: [KEEP/EXCLUDE]. Example: table: KEEP, washington: EXCLUDE, think: EXCLUDE, found: EXCLUDE. Words to analyze: \{List of words\}.''}

\subsection{Diagnostics (non-primary)}
\label{subsec:cdat-appedx-diagnostic}

\paragraph{Elbow distance.}
In general, elbow points on the Pareto front identify candidates that achieve the best balance between objectives before diminishing returns set in. The two task-agnostic baselines, Random and Common, maximize novelty and appropriateness, respectively, while minimizing the other, thereby defining the extreme ends of the trade-off. 

To quantify the elbow, we compute a signed Euclidean distance from each non-dominated model’s point to the line connecting the average scores of the Common and Random baselines using the formula:
\begin{equation*} 
d = \frac{ (y_2 - y_1)(x_0 - x_1) - (x_2 - x_1)(y_0 - y_1)}{\sqrt{(x_2 - x_1)^2 + (y_2 - y_1)^2}},
\end{equation*}
where $(x_0, y_0)$ is the model’s score, and $(x_1, y_1)$ and $(x_2, y_2)$ define the Common and Random baselines respectively. This score reflects the extent to which a model surpasses the linear trade-off between the most novel and the most appropriate solutions. A positive distance indicates that the point lies above the line, closer to the ideal region with maximized novelty and appropriateness, whereas a negative distance indicates that the point falls below it.

\paragraph{Distance to Human.}
\citet{heinen_semantic_2018} found that in a verb-generation task, when humans are cued to be creative, they implicitly enact both novelty and appropriateness criteria. As a qualitative, exploratory analysis, we use this implicit human definition of creativity as the ideal trade-off and judge models closer to this point as more creative.  

To obtain average human scores, 11 annotators (a convenience sample from the authors’ lab and personal networks) are recruited. All annotators reside in the Netherlands, hold a Master’s degree or higher, and include nine native Dutch speakers, one Korean speaker, and one Catalan–Spanish speaker. They rated a random subset of 73 cues (out of 550), yielding 104 annotations, of which 5 were excluded as invalid (total $n{=}99$). This left 70 cues with valid ratings: 29 had two independent ratings and 41 a single rating, balancing statistical efficiency and annotation cost. 

Inter-rater reliability was low for novelty (Krippendorff's $\alpha{=}-0.03$, $\bar{r}{=}-0.28$, $\mathrm{MAD}{=}5.20$) but higher for appropriateness ($\alpha{=}0.34$, $\bar{r}{=}0.44$, $\mathrm{MAD}{=}5.95$). Novelty scores were strongly rater-dependent (between-cue variance $\approx 47\%$, within-cue variance $\approx 53\%$), whereas appropriateness was more cue-dependent and consistent across raters (between-cue variance $\approx 67\%$, within-cue variance $\approx 33\%$).

\section{CDAT Results}
\label{sec:cdat-appdx-result}

\subsection{Counts and Sensitivity Data}
\label{subsec:cdat-appdx-counts}

Claude 3 Opus refuses to answer one cue for safety reasons (cue: \emph{prostitution}), and notably GPT-4’s answers to 68 out of 539 cues (12.62\%) at high temperature are rejected as invalid.

\subsection{2D Trade-off Visualizations}
\label{subsec:cdat-appdx-pareto-temp}

\begin{figure}[t]
  \centering
  \includegraphics[width=\linewidth, trim=0 0 0 18, clip]{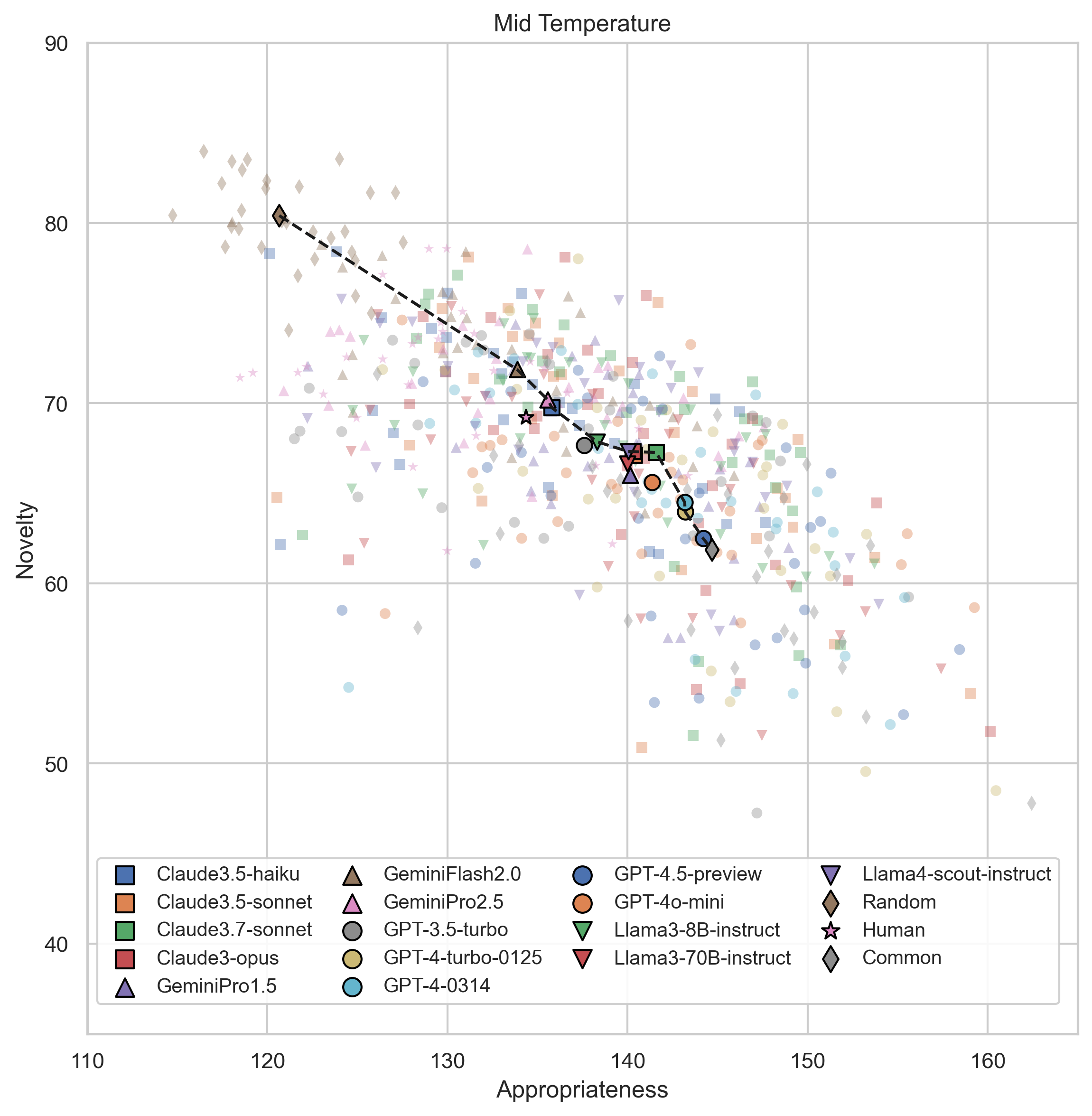}
  \caption{CDAT novelty–appropriateness landscape (middle temperature). Solid markers show model means; translucent markers show the first 30 per-cue points per model. The Pareto front (non-dominated models) traces optimal trade-offs. Means are jittered with Gaussian noise (mean $= 0$, SD $= 0.25$). All valid answers are used.}
  \label{fig:cdat-pareto-mid}
\end{figure}

\begin{figure}[t]
  \centering
  \includegraphics[width=\linewidth, trim=0 0 0 18, clip]{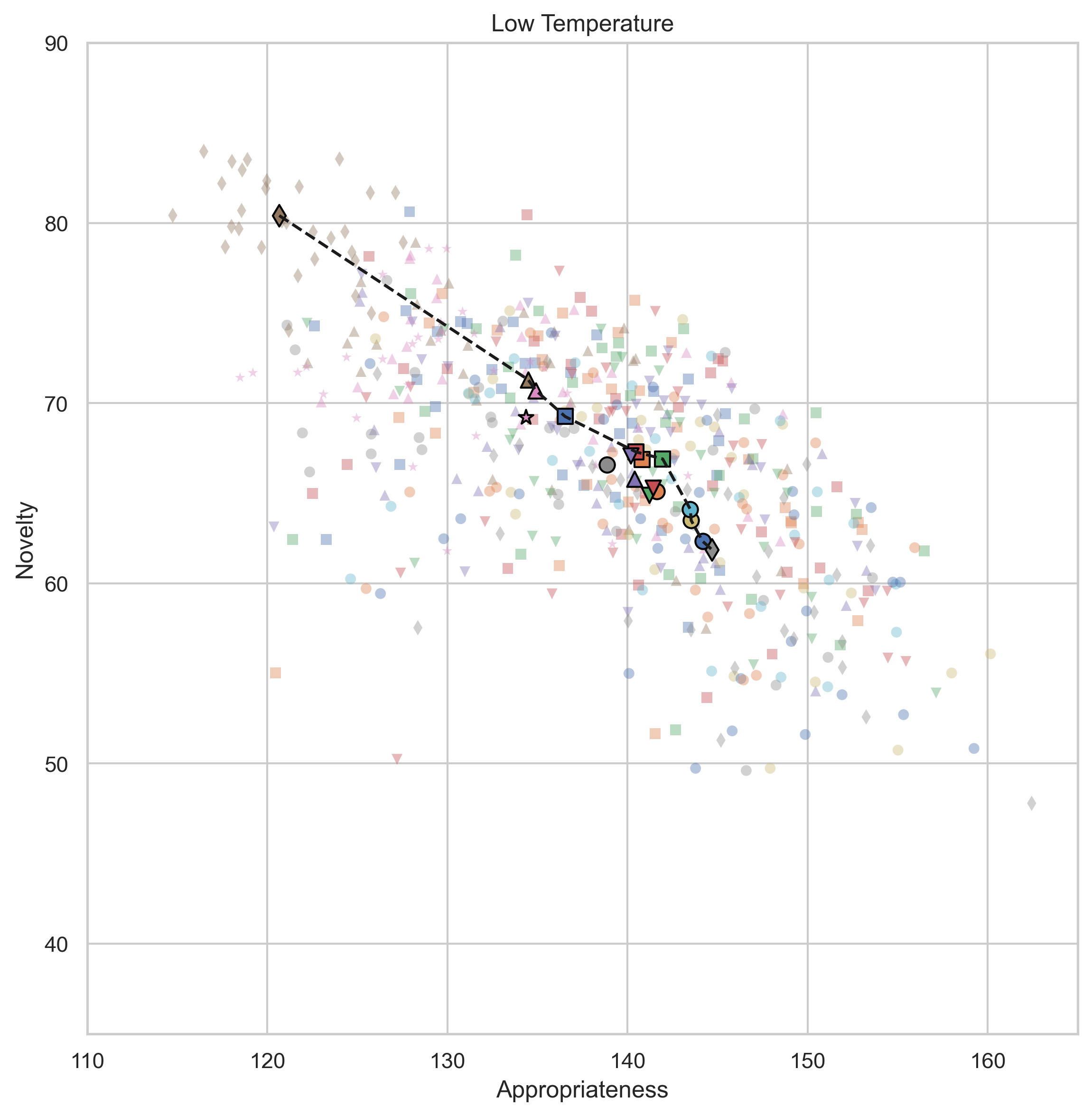}
  \caption{CDAT novelty–appropriateness landscape (low temperature).}
  \label{fig:cdat-pareto-low}
\end{figure}

\begin{figure}[t]
  \centering
  \includegraphics[width=\linewidth, trim=0 0 0 20, clip]{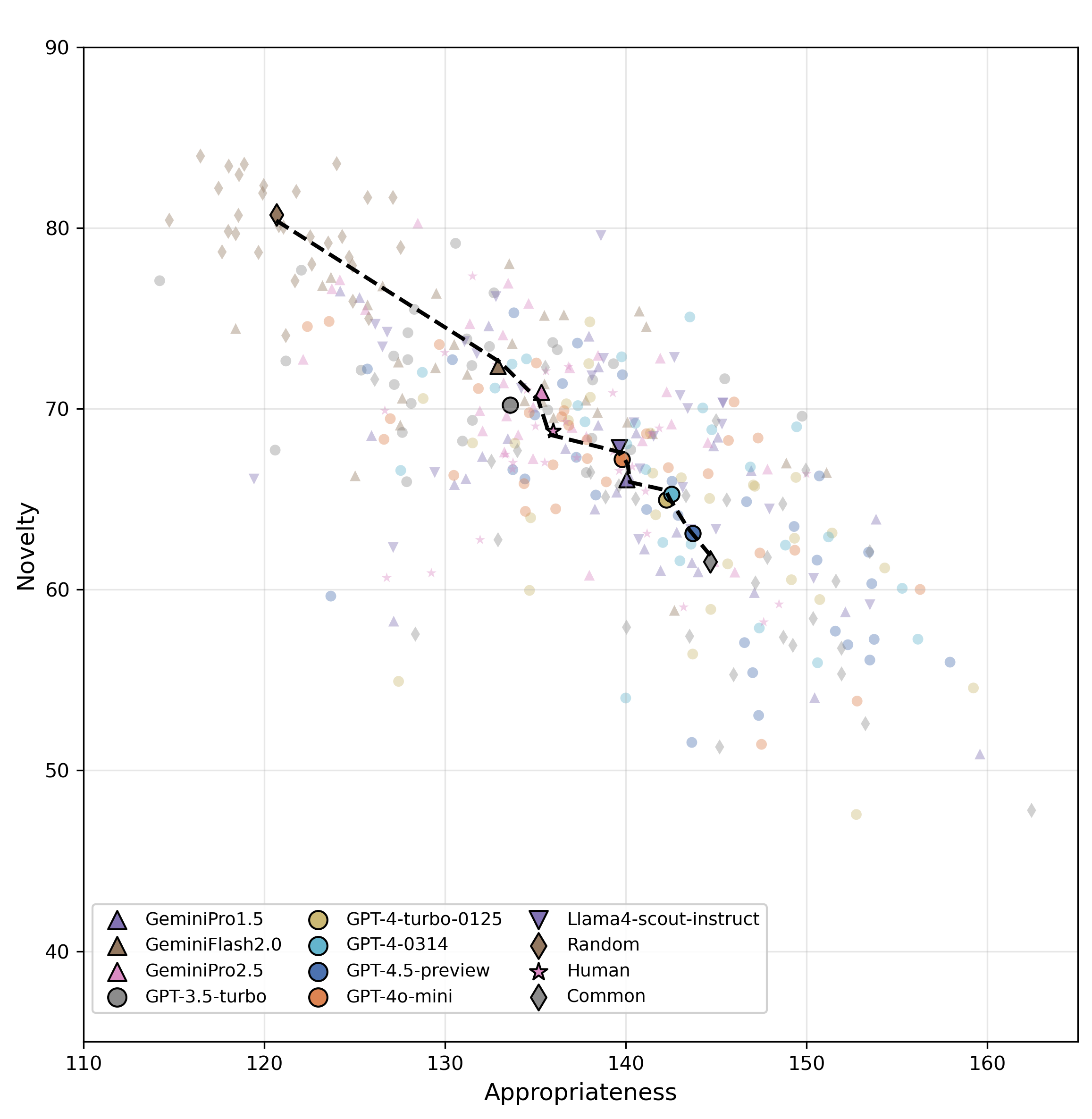}
  \caption{CDAT novelty–appropriateness landscape (high temperature).}
  \label{fig:cdat-pareto-high}
\end{figure}

We compare 2D novelty–appropriateness landscapes across temperatures to show that the general trend is consistent across these settings (Figures~\ref{fig:cdat-pareto-mid}--\ref{fig:cdat-pareto-high}). 

In addition to the main trend per family, a steeper line connecting the non-dominated models and more evident elbows suggests a stricter novelty–appropriateness trade-off and fewer efficient solutions at higher temperature, which aligns with previous findings suggesting that temperature has opposite relationships with novelty and appropriateness \citep{peeperkorn_is_2024}. 

\subsection{Scalar CDAT Statistics}
\label{subsec:cdat-appdx-novelty}

\begin{figure*}[t]
\centering
\includegraphics[width=\textwidth, trim=0 0 0 30, clip]{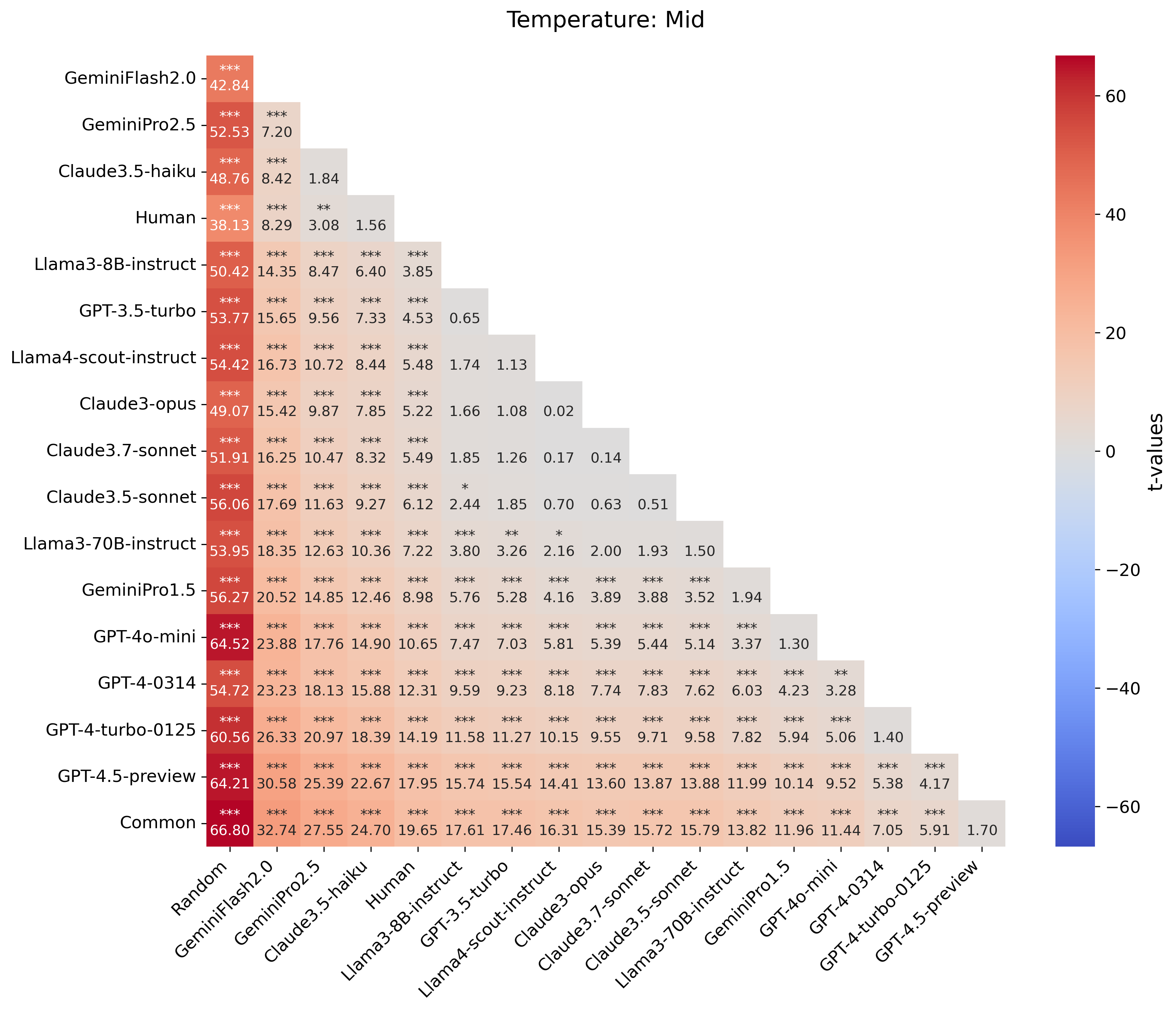}
\caption{(a) Pairwise t-value heatmap for models’ CDAT novelty scores (middle temperature). Each cell shows the contrast between the column model and row model (column model $-$ row model). Models are sorted by performance, with the top-performing model in the leftmost column. $*$: $p < .05$, $**$: $p < .01$, $***$: $p < .001$.}
\label{fig:cdat-novelty-mid}
\end{figure*}

\begin{figure*}[t]
\centering
\includegraphics[width=\textwidth, trim=0 0 0 30, clip]{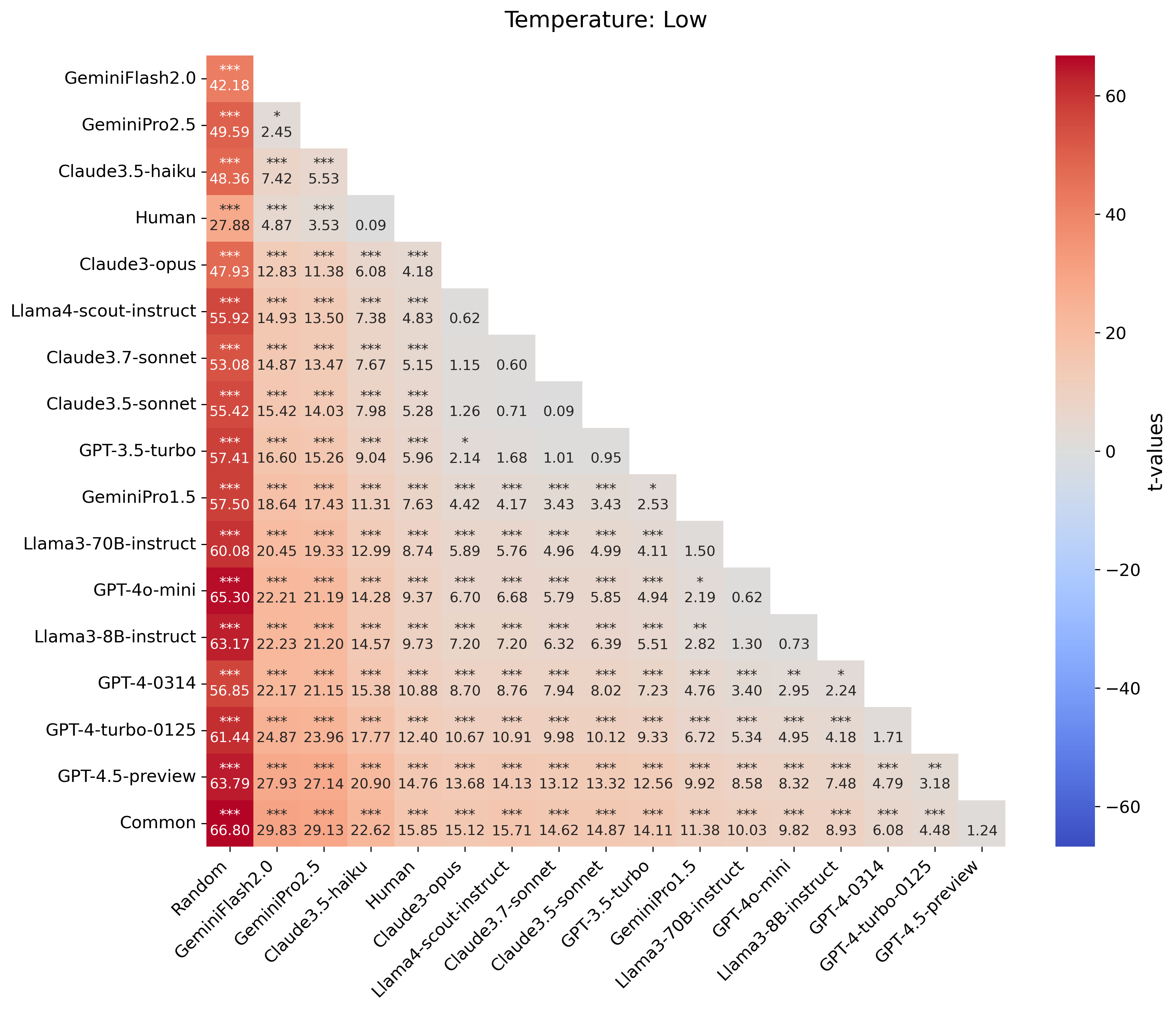}
\caption{(b) Pairwise t-value heatmap for models’ CDAT novelty scores (low temperature).}
\label{fig:cdat-novelty-low}
\end{figure*}

\begin{figure*}[t]
\centering
\includegraphics[width=\textwidth, trim=0 0 0 30, clip]{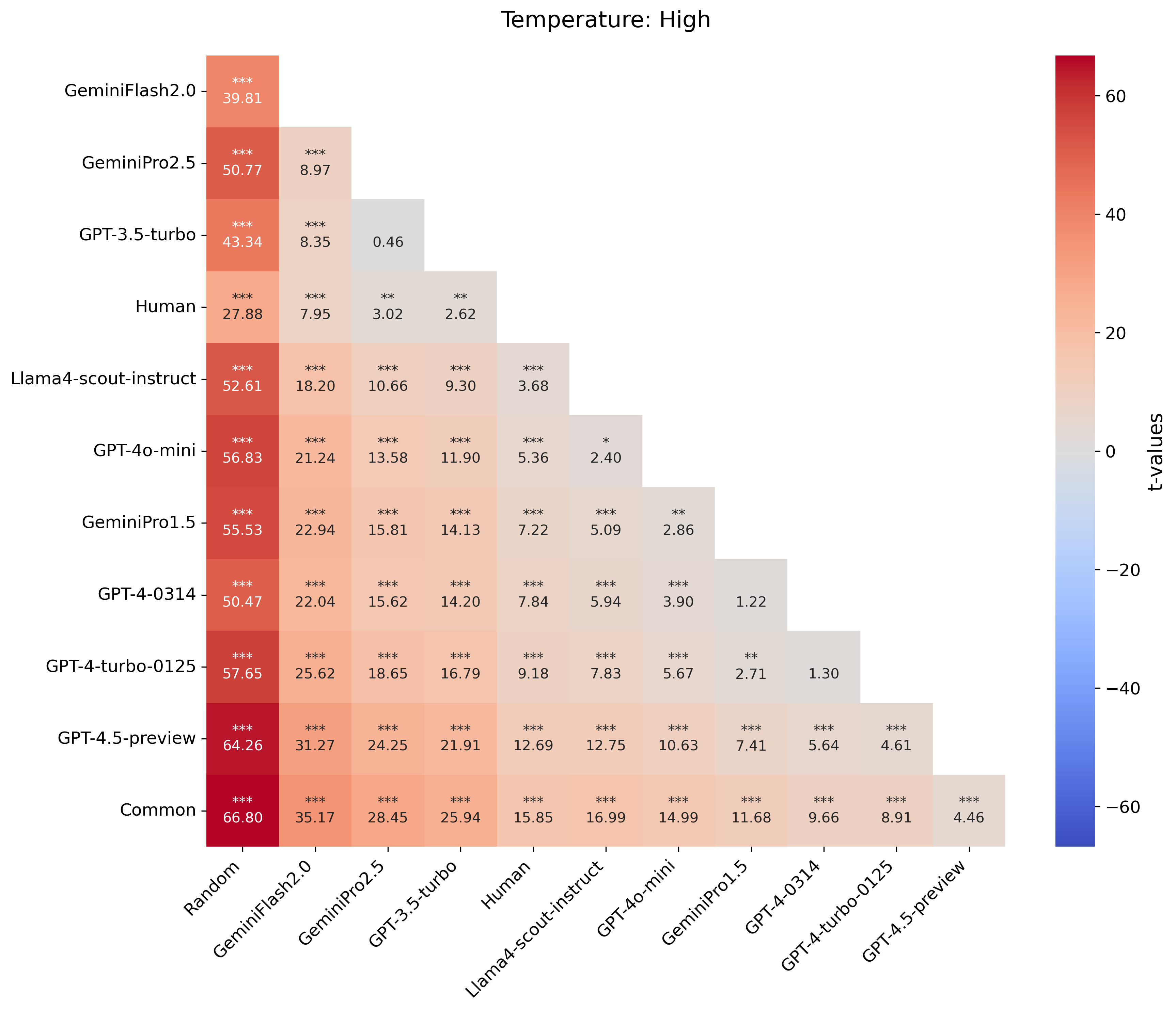}
\caption{(c) Pairwise t-value heatmap for models’ CDAT novelty scores (high temperature).}
\label{fig:cdat-novelty-high}
\end{figure*}

We conduct pairwise t-tests assessing differences in CDAT \emph{novelty} scores between models for each temperature condition (Figures~\ref{fig:cdat-novelty-mid}--\ref{fig:cdat-novelty-high}).

\subsection{Diagnostics (non-primary)}
\label{subsec:cdat-appdx-reslt-diagnostics}

\begin{figure}[t]
  \centering
  \includegraphics[width=\linewidth]{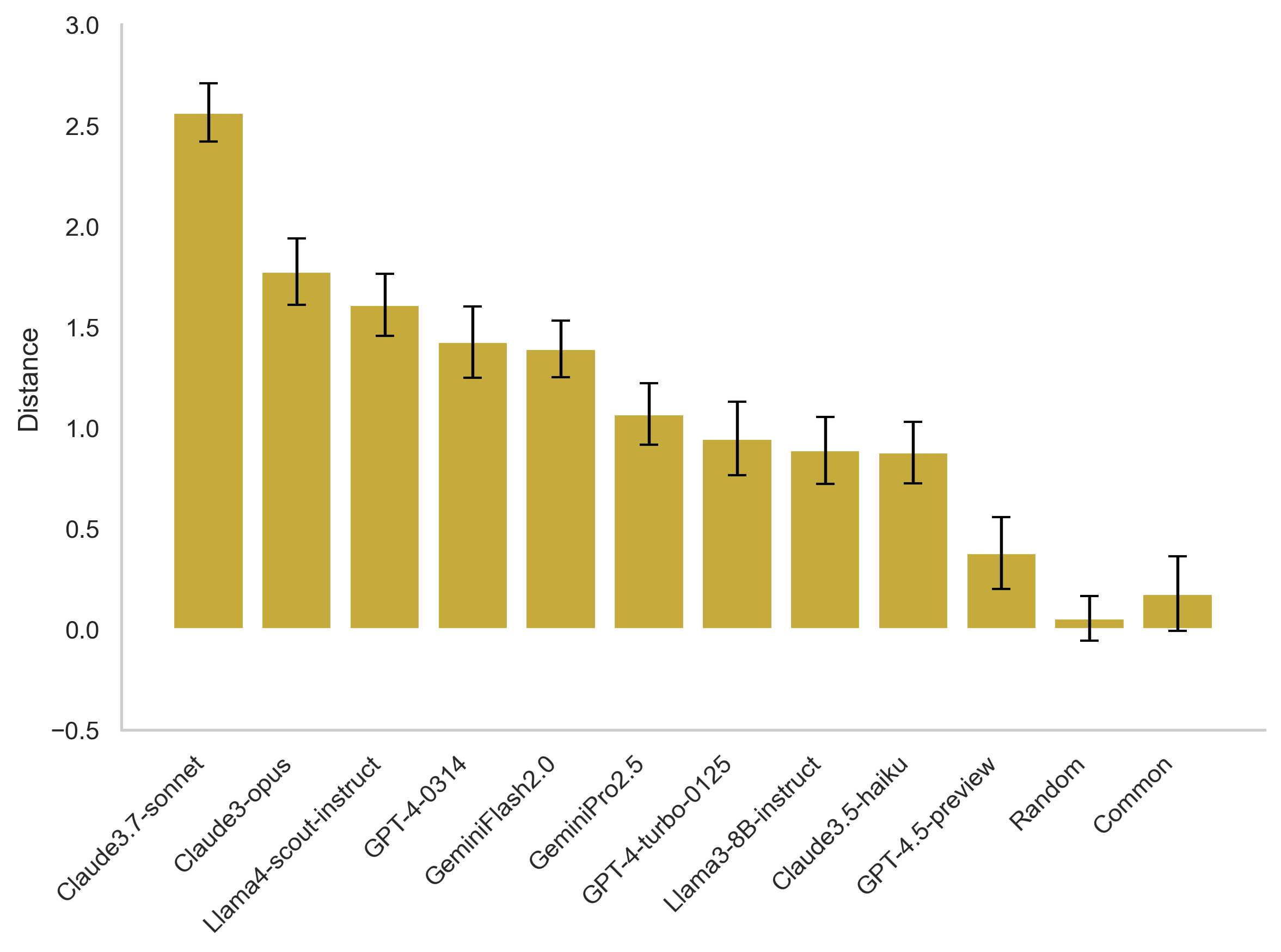}
  \caption{(a) The mean Elbow distance (middle temperature) with 95\% confidence intervals, including only non-dominated models.}
  \label{fig:cdat-elbow-mid}
\end{figure}

\begin{figure}[t]
  \centering
  \includegraphics[width=\linewidth]{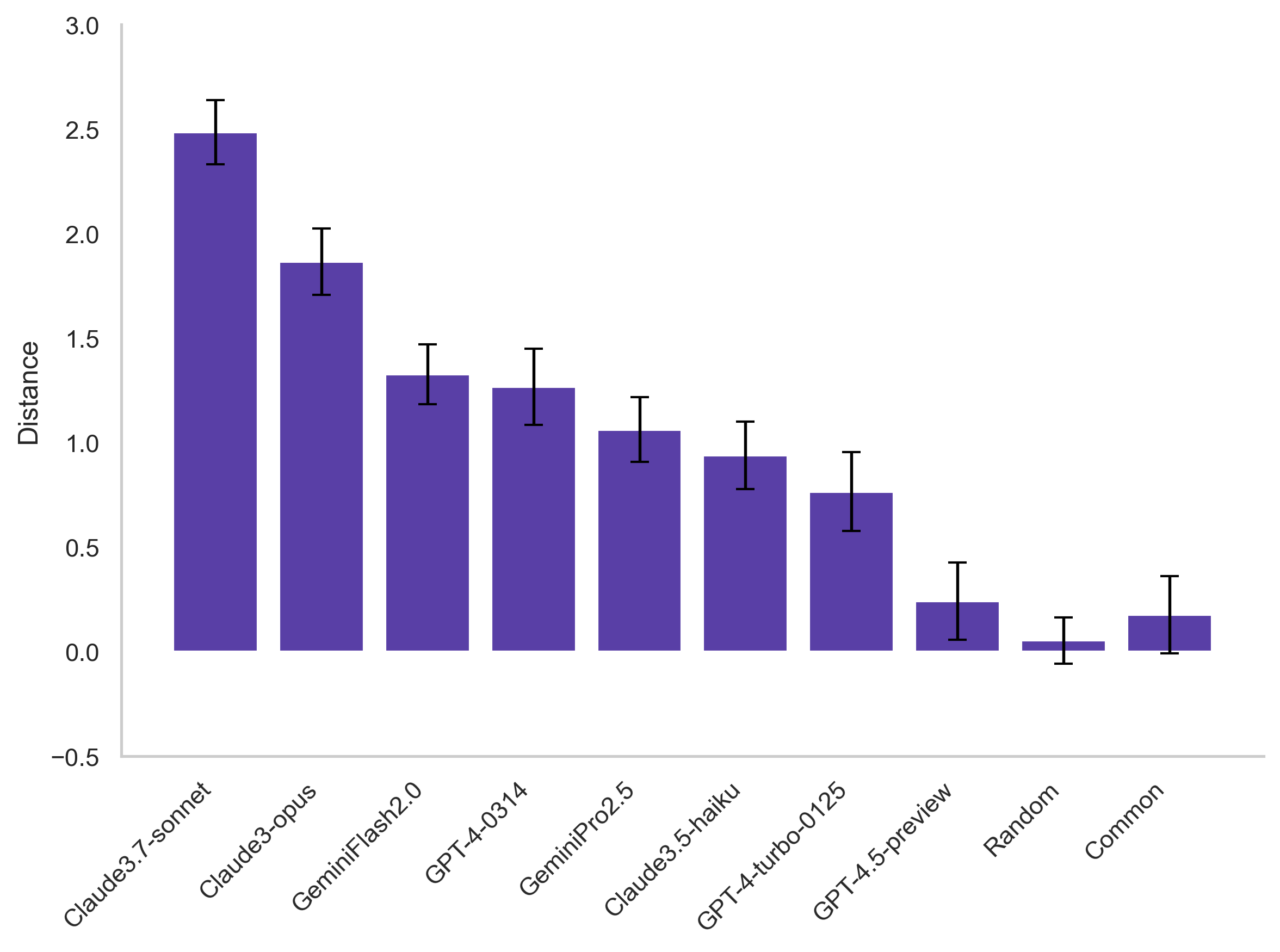}
  \caption{(b) The mean Elbow distance (low temperature).}
  \label{fig:cdat-elbow-low}
\end{figure}

\begin{figure}[t]
  \centering
  \includegraphics[width=\linewidth]{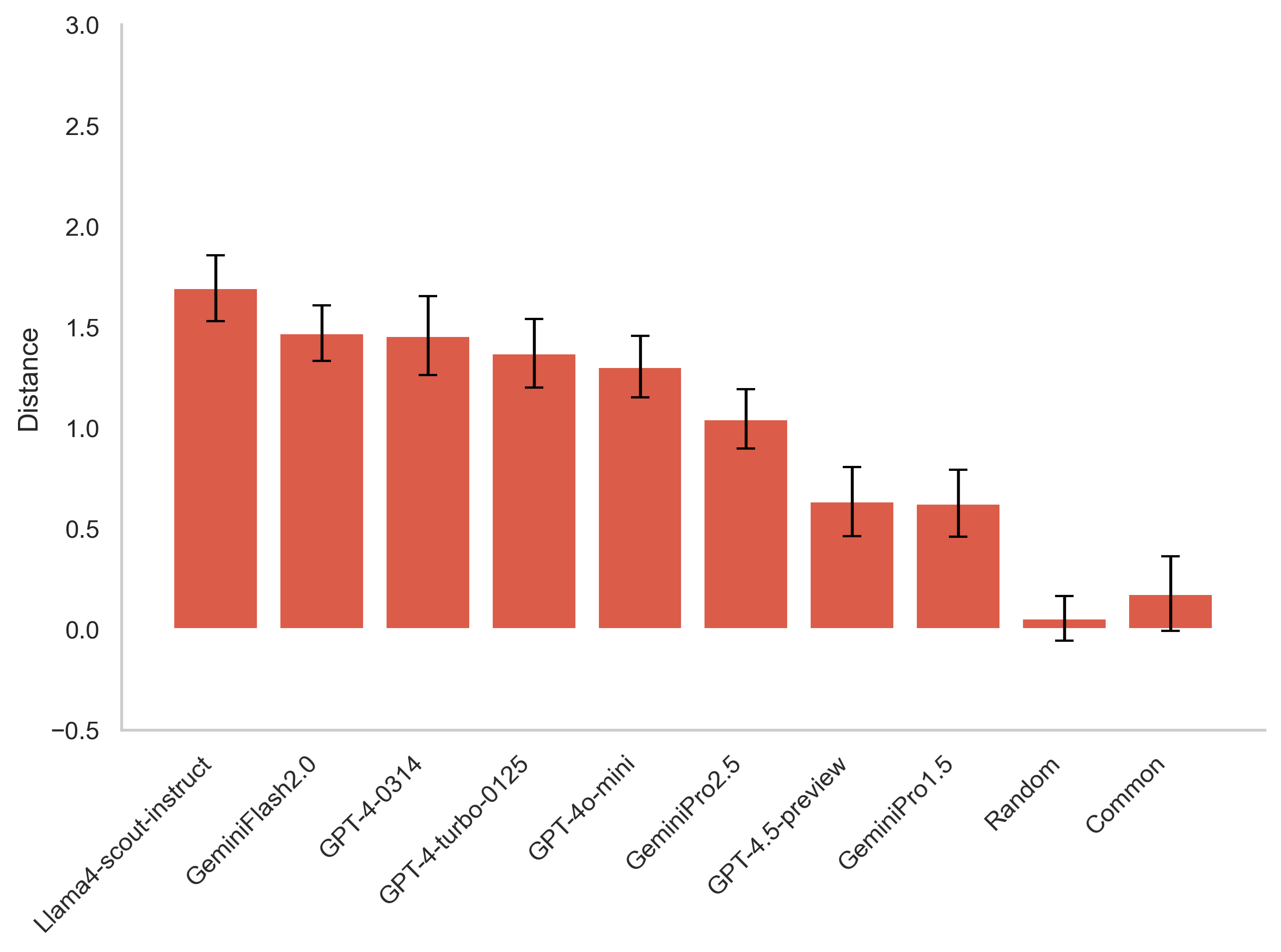}
  \caption{(c) The mean Elbow distance (high temperature).}
  \label{fig:cdat-elbow-high}
\end{figure}

\paragraph{Elbow distance.}
We compare Elbow distance values to identify models that exceed the trivial novelty–appropriateness trade-off, analyzing these values for each temperature setting (Figs.~\ref{fig:cdat-elbow-mid}--\ref{fig:cdat-elbow-high}).

\begin{figure*}[t]
\centering
\includegraphics[width=\textwidth]{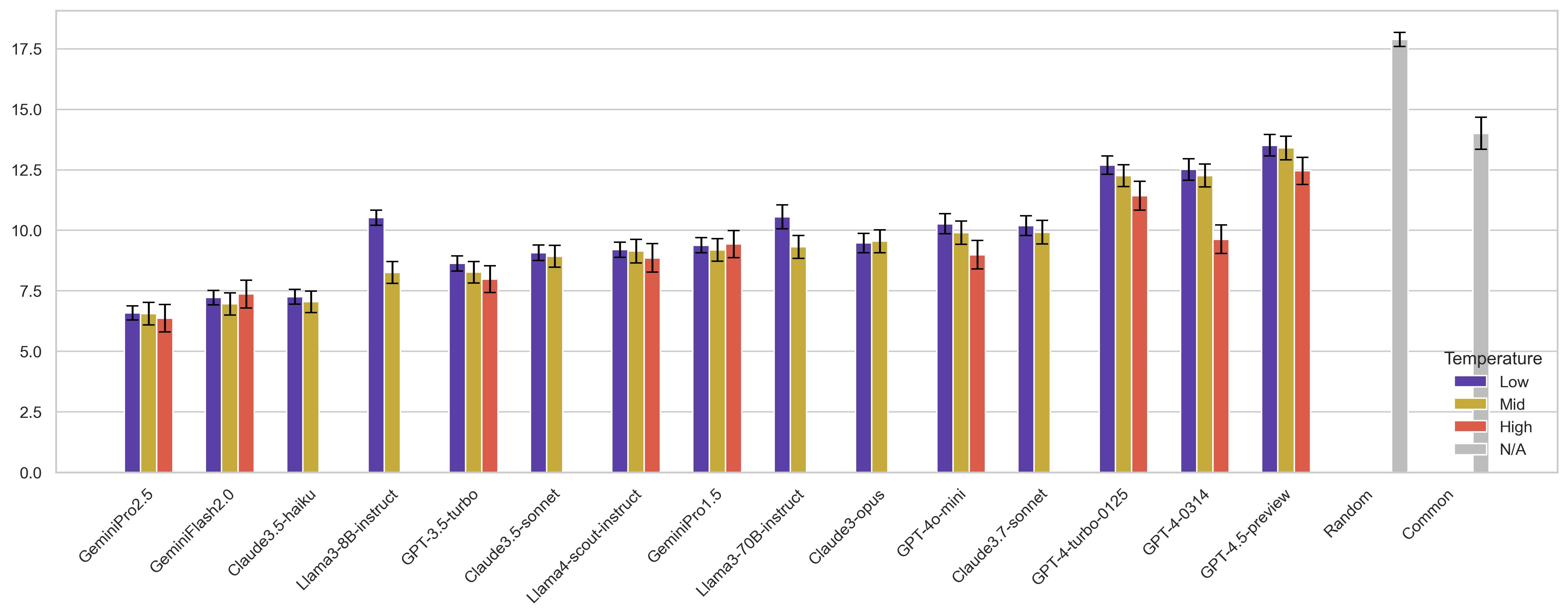}
\caption{The mean Euclidean distance from the human average score, sorted by top-performing model at the middle temperature. Note that some models do not support high temperature, and Random and Common baselines do not vary by temperature. Error bars show 95\% confidence intervals.}
\label{fig:cdat-human-avg}
\end{figure*}

\paragraph{Distance to Human.}
We compare Euclidean distances to the human average at each temperature to identify models that align more closely with humans’ implicit novelty–appropriateness balance (see Figure~\ref{fig:cdat-human-avg}). Model-specific distances to the human average, together with their significance levels across temperatures, are available in our GitHub repository.

\section{CDAT Additional Analyses}
\label{sec:cdat-appdx-additional}

\subsection{Interpreting the Trade-off}
\label{subsec:cdat-appdx-ablation}

\begin{figure*}[t]
\centering
\includegraphics[width=\textwidth]{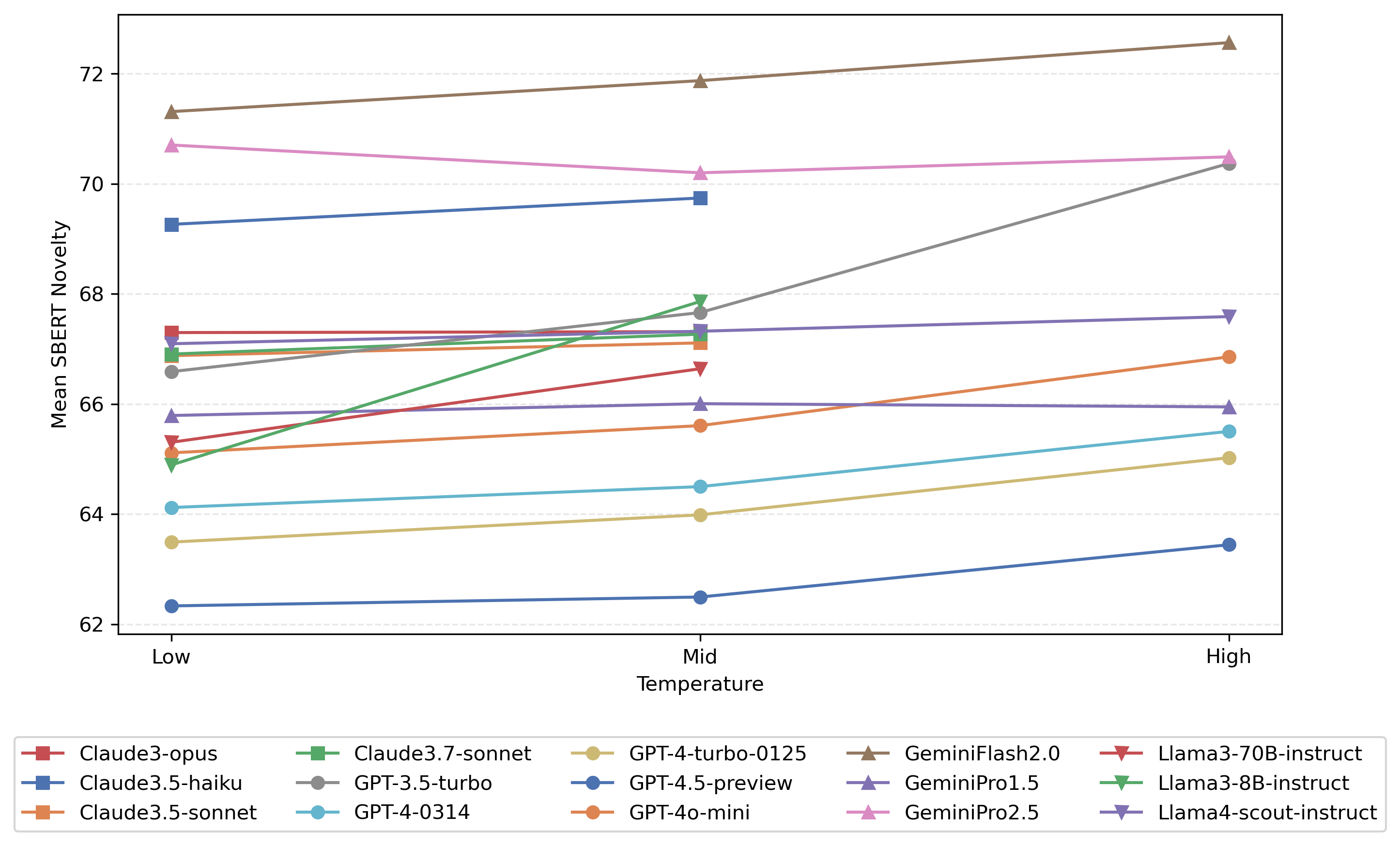}
\caption{Transition of mean SBERT-based novelty scores for each model across temperatures.}
\label{fig:cdat-appdx-delta-all}
\end{figure*}

\paragraph{Temperature-based diagnostic of novelty constraints.}
As an initial step toward separating the effects of instruction and safety alignment from models' underlying creative capacity, we examine how mean novelty scores change as sampling temperature increases. If novelty remains low for larger or more aligned models even under higher-temperature decoding, this suggests that reduced novelty may not be solely attributable to conservative sampling, and post-training alignment might have induced a form of mode collapse.

Figure~\ref{fig:cdat-appdx-delta-all} shows the transition of mean SBERT-based novelty scores across temperatures. For several highly aligned models, such as Claude~3~Opus, Gemini~Pro~2.5, and Llama~4~Scout, novelty scores remain relatively flat as temperature increases. In contrast, less aligned or smaller models, such as Gemini~Flash~2.0, Claude~3.5~Haiku, Llama~3.1~8B~Instruct, GPT-4o~mini, exhibit larger gains in novelty. This pattern is consistent with the hypothesis that post-training alignment constrains the accessible output distribution, potentially inducing a form of conservative generation or mode restriction.

This analysis is correlational rather than causal; temperature-based probing cannot disentangle alignment effects from architectural, training-data, or representation-level factors, and similar flat patterns could arise from multiple mechanisms. Confirming an alignment effect would require controlled interventions that explicitly manipulate alignment strength.

\begin{figure*}[t]
\centering
\includegraphics[width=\textwidth]{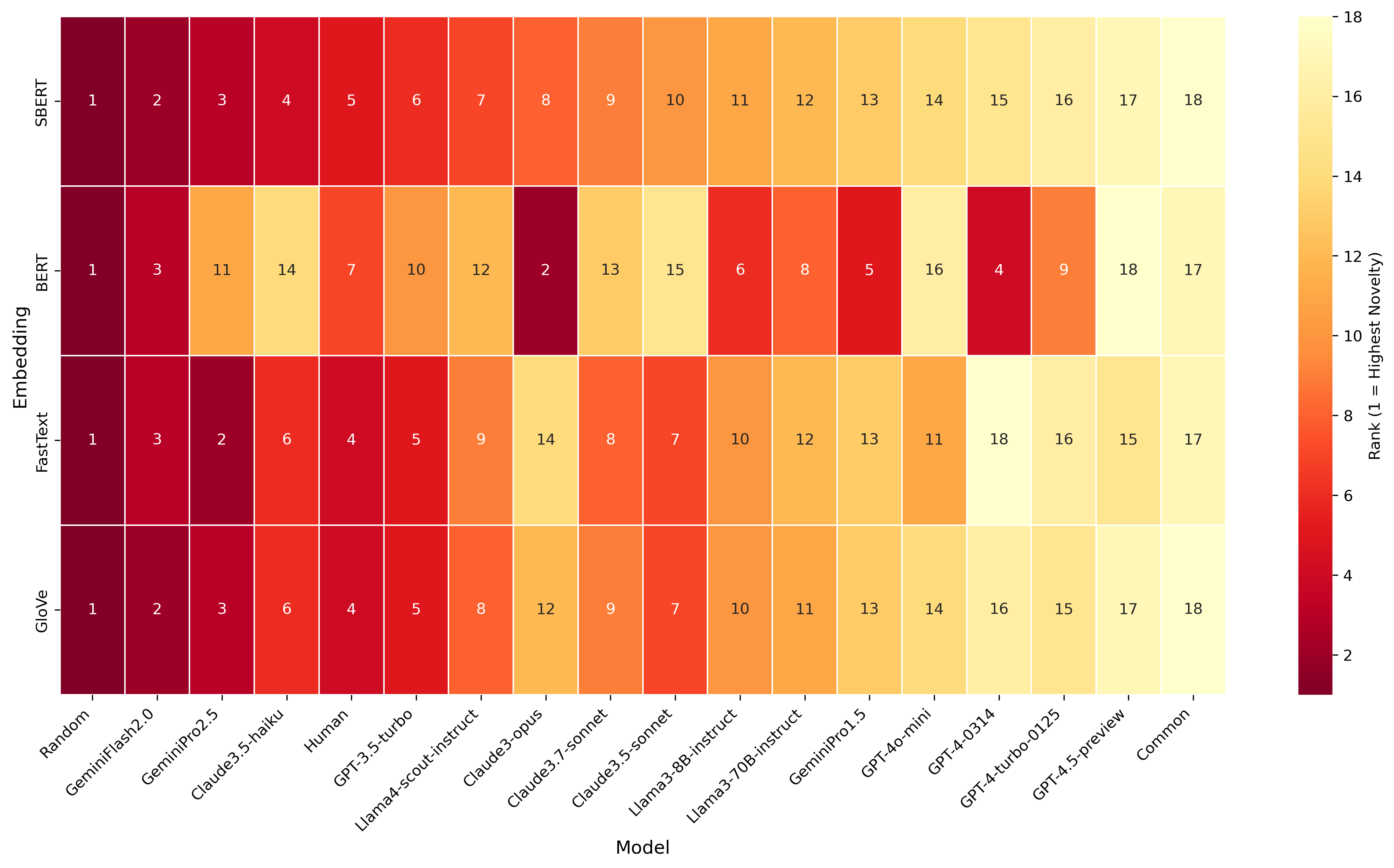}
\caption{CDAT novelty rankings across four embedding models (SBERT, BERT, fastText, GloVe). BERT shows the average of layers 3–9. Darker colors denote higher novelty (lower rank numbers). Models are sorted by SBERT ranking. Similar colors across embedding types indicate similar rankings.}
\label{fig:cdat-appdx-embedding-ranking}
\end{figure*}

\paragraph{Prompting and decoding as capability probes.}
We used a single, uniform prompt per condition to avoid per-model tuning. Because decoding and prompt interpretation can reweight novelty vs.\ appropriateness, an automated, bounded prompt-search protocol could estimate an upper bound on each model’s CDAT potential while keeping the search procedure identical across families. Likewise, controlled studies that factor temperature, nucleus/top-$k$ sampling, and length penalties can quantify how much of the trade-off is attributable to decoding rather than model internals.

\paragraph{Mechanism tests.}
To more directly test the frontier-shift hypothesis suggested by the above analyses, future work should conduct controlled ablations on open-weight models that vary one factor at a time, including pretraining data diversity, instruction tuning strength, refusal policies, and other guardrail objectives. Comparing base and instruction-tuned checkpoints under matched decoding settings would allow measurement of induced displacements in the two-dimensional novelty–appropriateness plane. A complementary, data-centric intervention would suppress high-frequency associative continuations during training or decoding to test whether models can recover novelty without sacrificing appropriateness.

\subsection{Measurement Choices and Their Implications}
\label{subsec:cdat-embedding-power}

\paragraph{Power analysis.}
To assess the sample size that is sufficient to reliably detect the observed effects used in the appropriateness gate, we conduct a post-hoc power analysis. To achieve $80\%$ power at $\alpha{=}0.001$, approximately six responses would suffice for the Random vs.\ Human comparison (Cohen's $d{=}3.59$), about nine responses for the Random vs.\ Gemini 2.0 Flash comparison, the smallest-effect model at both low (Cohen's $d{=}2.54$) and middle (Cohen's $d{=}2.46$) temperatures, and about 11 responses for Random vs.\ GPT-3.5 Turbo, the smallest-effect model at the high temperature (Cohen's $d{=}2.07$). Our actual sample ($n{=}550$ responses) therefore provides more than sufficient power to detect these effects.

\paragraph{Embedding sensitivity analysis.}
\label{subsec:cdat-appdx-embedding-sensitivity}

We assess the robustness of the CDAT metric to the choice of embedding representation via a sensitivity analysis across four models: GloVe, fastText,\footnote{Model: \texttt{fastText Common Crawl English (cc.en.300.bin)}} BERT,\footnote{Model ID: \texttt{bert-base-uncased}} and SBERT. For each model, we compute CDAT novelty scores for the LLMs (Figure~\ref{fig:cdat-appdx-embedding-ranking}) and then calculate Spearman rank correlations among the resulting rankings.

For BERT, we obtain word embeddings by mean-pooling all token embeddings ([CLS], word tokens, [SEP]) within each of layers 3–9; mean pooling over middle layers has been reported as effective for semantic similarity \citep{reimers_sentence-bert_2019, ma_universal_2019, rogers_primer_2020}. Novelty rankings from GloVe, fastText, and SBERT are strongly and significantly aligned (pairwise Spearman’s $\rho$ between 0.87 and 0.96; all $p{<}0.001$). By contrast, BERT-based rankings with any layers are not significantly correlated with any of the other embeddings. We therefore average the seven layer-wise pooled BERT vectors (layers 3–9) and report results using this final BERT-averaged embedding (fastText–BERT: $\rho{=}-0.07$, $p{=} 0.81$; GloVe–BERT: $\rho{=}0.14$, $p{=}0.62$; SBERT–BERT: $\rho{=}0.28$, $p{=}0.29$). 

A likely reason BERT yields different rankings is the anisotropic geometry of its embedding space \citep{ethayarajh_how_2019}: BERT representations concentrate in a few dominant directions, which may cause cosine similarities to behave differently than in more isotropic embeddings such as GloVe, fastText, or SBERT.
Post-processing, such as static embeddings distilled from contextualized representations, can provide BERT with high-quality type-level word representations \citep{ethayarajh_how_2019}.

Note about the GloVe model: in addition to general rejections by the other embedding models, one answer by GPT-3.5~Turbo and GPT-4~mini at high temperature, two answers by GPT-4 and GPT-4~Turbo at high temperature, one answer by Llama~4~Scout at low temperature, and 12 responses from the Random baseline are rejected because fewer than seven generated words exist in the GloVe model. All cue words are present in GloVe. 

\subsection{Scope Extensions and Open Directions}
\label{subsec:cdat-appdx-scope}

\paragraph{Human vs.\ machine-native metrics.}
We view CDAT as a complementary, human-grounded part of the evaluation space: it targets a well-studied creativity construct in a minimal, controlled setting, while machine-native metrics can capture other model-specific aspects of generativity. Although future work may directly optimize LLMs for statistical novelty, CDAT helps identify a clear trend: larger LLMs are often post-trained in ways that reduce their ability to produce statistical novelty.

\paragraph{Scope extensions.}
Three directions extend the framework without sacrificing clarity. (i) \emph{Cue design:} move beyond single nouns to relational or compositional cues (e.g., two-word cues or typed relations) to test whether models can diversify around more structured contexts. (ii) \emph{Linguistic coverage:} evaluate cross-lingual CDAT with language-appropriate lexical resources and embedding spaces, assessing whether trade-offs replicate across morphologies and scripts. (iii) \emph{Multimodality:} adapt CDAT to image–text settings by conditioning on visual cues and scoring textual novelty subject to visual appropriateness using vision–language embeddings, keeping the gate-and-score logic intact.

\end{document}